\documentclass[10pt,twocolumn,letterpaper]{article}

\usepackage[pagenumbers]{cvpr} % To force page numbers, e.g. for an arXiv version

\usepackage{graphicx}
\usepackage{amsmath}
\usepackage{amssymb}
\usepackage{booktabs}
\usepackage{breakcites}
\usepackage[pagebackref,breaklinks,colorlinks]{hyperref}
\usepackage[numbers,sort]{natbib} % sort citations
\usepackage[toc,page]{appendix}

\usepackage[capitalize]{cleveref}
\crefname{section}{Sec.}{Secs.}
\Crefname{section}{Section}{Sections}
\Crefname{table}{Table}{Tables}
\crefname{table}{Tab.}{Tabs.}

\def\cvprPaperID{6430} % *** Enter the CVPR Paper ID here
\def\confName{CVPR}
\def\confYear{2022}

\usepackage[table]{xcolor}
\usepackage{cuted}
\usepackage{xcolor}
\usepackage{pifont}
\usepackage{lipsum}
\usepackage{wrapfig}
\usepackage{overpic}
\usepackage{overpic}          %< add raw math symbols to figures
\usepackage{capt-of}
\usepackage{multirow}
\usepackage{booktabs}
\usepackage{comment}
\usepackage{comment}
\usepackage{amsmath}
\usepackage{amssymb}
\usepackage{adjustbox}
\usepackage{soul,color}
\usepackage{enumitem} %< control spacing in itemize/enumerate/...
\usepackage[accsupp]{axessibility}
\definecolor{turquoise}{cmyk}{0.65,0,0.1,0.3}
\definecolor{purple}{rgb}{0.65,0,0.65}
\definecolor{dark_green}{rgb}{0, 0.5, 0}
\definecolor{orange}{rgb}{0.8, 0.6, 0.2}
\definecolor{red}{rgb}{0.8, 0.2, 0.2}
\definecolor{darkred}{rgb}{0.6, 0.1, 0.05}
\definecolor{blueish}{rgb}{0.0, 0.3, .6}
\definecolor{light_gray}{rgb}{0.7, 0.7, .7}
\definecolor{pink}{rgb}{1, 0, 1}
\definecolor{greyblue}{rgb}{0.25, 0.25, 1}

\usepackage{blindtext}

\renewcommand{\paragraph}[1]{\vspace{1em}\noindent\textbf{#1}.}

\newcommand{\xmark}{\ding{55}}%

\newcommand{\weidi}[1]{\textcolor{magenta}{[Weidi: #1]}}

\newcommand{\best}[1]{{\color[HTML]{3531FF}{\textbf{#1}}}}
\newcommand{\sbest}[1]{{\color[HTML]{FE0000}{\textit{#1}}}}
\newcommand{\mailtodomain}[1]{\href{mailto:#1@robots.ox.ac.uk}{\nolinkurl{#1}}}

\begin{document}
% !TEX root = ../main.tex
\title{{Label, Verify, Correct: A Simple Few Shot Object Detection Method} \\
\vspace{-12pt}}

\author{Prannay Kaul\textsuperscript{1} \qquad Weidi Xie\textsuperscript{1,2} \qquad Andrew Zisserman\textsuperscript{1}\\
\textsuperscript{1}Visual Geometry Group, University of Oxford \qquad \textsuperscript{2}Shanghai Jiao Tong University\\
%{\tt\small \{\mailtodomain{prannay}, \mailtodomain{weidi}, \mailtodomain{az}\}@robots.ox.ac.uk}\\
{\tt\small \href{http://www.robots.ox.ac.uk/~vgg/research/lvc/}{\url{http://www.robots.ox.ac.uk/~vgg/research/lvc/}}}
% \and
\vspace{-12pt}
}

% !TEX root = ../main.tex
\twocolumn[{
\renewcommand\twocolumn[1][]{#1}%
\maketitle

% \vspace{-10pt}
\begin{center}
\scriptsize
\includegraphics[width=1.0\linewidth]{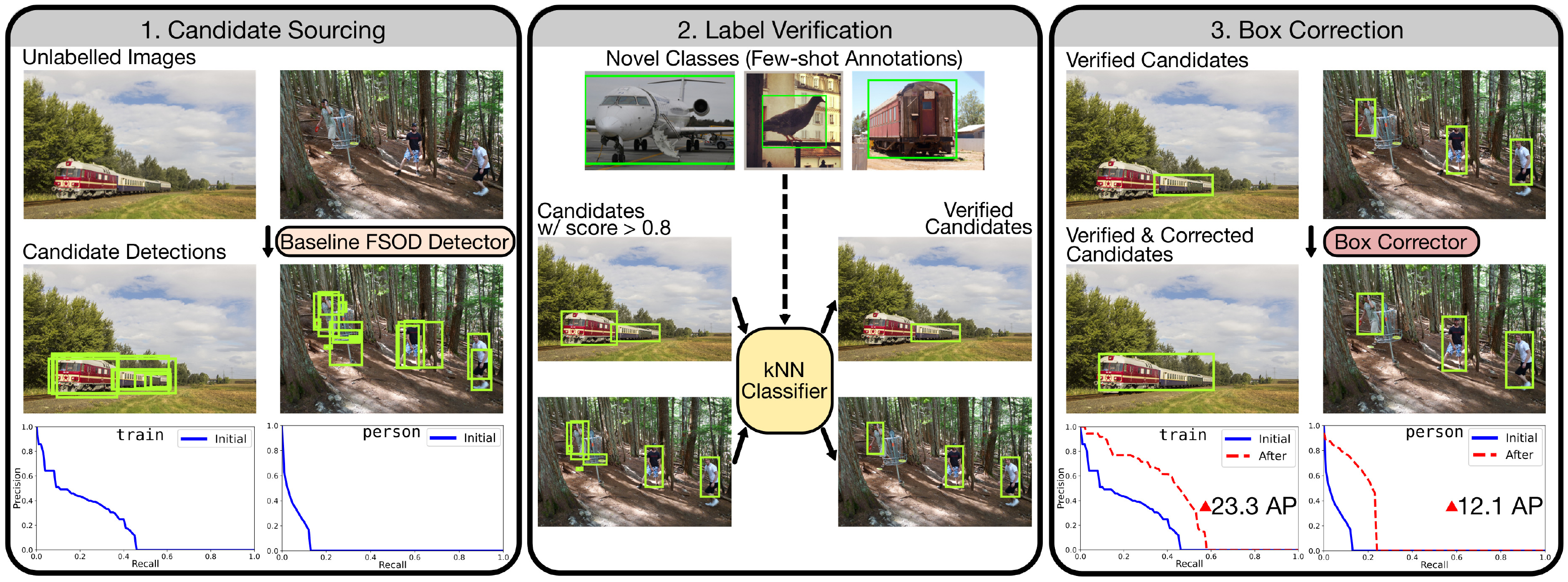} %eps
\vspace{-10pt}
\captionof{figure}{{\bf Label, Verify, Correct:}
Object detectors na\"ively trained with $K$-shot annotations perform poorly on novel classes (bottom left).
We propose to expand the novel class annotation set and re-train {\em end-to-end}.
1. Given a baseline few-shot object detector, noisy candidate detections are sourced from unlabelled images (left).
2. Labels for each candidate detection are verified by a $k$NN classifier, constructed from a self-supervised model using the same few-shot annotations,
removing large numbers of false positives (centre).
3. A specialised box corrector, drastically improves the remaining bounding boxes, yielding high-quality pseudo-annotations (right).
Re-training end-to-end with our pseudo-labelling method yields a large performance boost on novel class detection, improving precision and recall (bottom right).}
\label{fig:pipeline}
\end{center}
}]

\maketitle
% !TEX root = ../main.tex

\begin{abstract}
\vspace{-10pt}
The objective of this paper is few-shot object detection (FSOD) --
the task of expanding an object detector for a new category given only a few instances for training.
We introduce a simple pseudo-labelling method to source high-quality pseudo-annotations from the training set, for each new category,
vastly increasing the number of training instances and reducing class imbalance;
our method finds previously unlabelled instances.

Na\"ively training with model predictions yields sub-optimal performance;
we present two novel methods to improve the precision of the pseudo-labelling process:
first, we introduce a verification technique to remove candidate detections with incorrect class labels;
second, we train a specialised model to correct poor quality bounding boxes.
After these two novel steps,
we obtain a large set of high-quality pseudo-annotations that allow our final detector to be trained end-to-end.
Additionally, we demonstrate our method maintains base class performance, and the utility of simple augmentations in FSOD.
While benchmarking on PASCAL VOC and MS-COCO,
our method achieves state-of-the-art or second-best performance compared to existing approaches across all number of shots.
\end{abstract}

\begin{comment}
\begin{abstract}
The objective of this paper is few-shot object detection (FSOD) --
the task of learning~\weidi{or use `expanding'?} an object detector for a new category given only a few instances as training.
This is a challenging task, and we identify two reasons why previous approaches have underperformed -- class imbalance and ``supervision-collapse'',
which leads to overfitting.

To address these issues, we introduce a simple pseudo-labelling method to source high-quality pseudo-annotations, from the training set, for each new category;
vastly increasing the number of training instances and reducing class imbalance.
Na\"ively training with the model's predictions yields sub-optimal performance,
we present two novel methods to improve the precision of such pseudo-labelling process:
First, we introduce a verification technique to remove the sourced candidates with incorrect class labels.
Second, we train a specialised model to correct poor quality bounding boxes.
After these two-step filtering,
we end up with a large set of high-quality pseudo-annotations that allow our final detector to be trained end-to-end.
Additionally, we demonstrate our method maintains base class performance and the utility of simple augmentations in FSOD.
While benchmarking on PASCAL VOC and MS-COCO, we show comparable or state-of-the-art performance to existing approaches across all number of shots.
\end{abstract}
\end{comment}

% !TEX root = ../main.tex
\vspace{-15pt}
\section{Introduction}
\label{sec:introduction}
\vspace{-5pt}
Object detection refers to the task of determining if an image contains objects of a particular category,
and if so, then localising them.
In recent years,
the community has seen tremendous successes in object detection by training computational models for a set of
pre-defined object classes~\cite{Ren16,Lin2017a,Tan2020,Xu2021,Dai2021,Qiao2021a,Han2021a},
with large numbers of human annotated labels, \eg~ MS-COCO~\cite{Lin14}, and PASCAL VOC~\cite{Everingham15}.
However, such training paradigms have limited the model to only perform well on a closed,
small set of categories for which large training data is available.

In contrast, humans can continuously expand their vocabularies,
learning to detect a much larger set of categories,
even with access to only a few examples~\cite{Smith2017}.
This is also a  desirable ability for modern computer vision systems and is studied in the
task of few-shot object detection~(FSOD)~\cite{Kang2019,Yan2019,Fan2021,Wu2020a,Sun2021}.
The goal of our work is FSOD: given an existing object detector that
has been trained on abundant data for some categories, termed \emph{base} categories,
we wish to learn to detect {\em novel} categories using only a few annotations, \eg~1--30 per category,
whilst maintaining performance on the original base categories.

In this paper,
we carefully examine the training procedure of the Faster R-CNN two-stage detector,
and identify two critical factors that limit its performance on FSOD.
{\em First}, training on base categories leads to a form of ``supervision collapse''~\cite{Doersch2020};
this model is naturally trained against detecting instances from novel classes,
as they were unlabelled and treated as background; % or `false positive' with respect to base classes; \prannay{??}
{\em second},
the FSOD problem involves learning from extremely unbalanced data --
only $K$ instances~($K \leq 30$) are available per novel class for training,
so the number of training samples for {\em base} categories is far larger than that for {\em novel} ones.
A model that overfits to a small number~($K$) of novel instances naturally
lacks generalisation ability.

We adopt a simple {\em pseudo-labelling} technique (see Figure~\ref{fig:pipeline}) and address both the factors that limit performance:
we show that the Region Proposal Network (RPN) can be modified to successfully propose regions for the novel categories and
use a detector trained with the few-shot {\em novel} data to label these regions over the images
of the large training dataset, producing a set of candidate detections for each novel category.
The novelty of our approach is in the two steps used to improve the precision of this candidate set:
{\em first}, we build a classifier for novel categories to verify the candidate detections using
features from a network trained with self-supervision (see Figure~\ref{fig:pipeline}, centre); % on the training dataset;
{\em second}, we train a specialised box regressor that improves the quality of the bounding box of the verified
candidates (see Figure~\ref{fig:pipeline}, right).
% The first step avoids the `supervision collapse' of the original features trained on the base categories,
The two steps together yield a large set of high precision pseudo-annotations for the novel categories,
removing the class imbalance in the training data.
This enables the detector to be trained end-to-end using the pseudo-annotations for the novel categories together with the original groundtruth annotations for base categories,
and simultaneously avoids the impact of ``supervision collapse'' for {\em all} detector features.

To summarise, our contributions are as follows:
(i) we carefully examine the problem of few-shot object detection with the modern two-stage object detector, {\em e.g.}~Faster R-CNN,
and identify the issue of ``supervision collapse'';
(ii) we introduce a novel verification and correction procedure to pseudo-labelling,
which significantly improves the precision of pseudo-annotations, both class labels and bounding box coordinates;
(iii) we analyse several critical components of data augmentation and conduct thorough ablation studies to validate their necessity;
(iv) with the combination of pseudo-labelling and aggressive data augmentations,
we set state-of-the-art~(SotA) or comparable performance using a standard Faster R-CNN,
for both the challenging MS-COCO benchmark and the PASCAL VOC benchmark.

We discuss limitations of our work and potential ethical concerns in Appendix~\ref{app:limitations} and ~\ref{app:ethical_concerns}, respectively.
Code and pre-trained models are available from the project webpage.

% \weidi{we need to make a decision on the terminology, do we want to use query expansion, or simple self-training with pseudo-annotations.}

% !TEX root = ../main.tex
\vspace{-5pt}
\section{Related Work}
\label{sec:related_work}
\vspace{-5pt}
\par{\noindent \bf Object Detection}
is one of the classical problems in computer vision,
which makes it impossible to present a full overview here.
We therefore only outline some key milestones here.
In general, recent object detection methods can be cast into two sets:
one-stage and two-stage detectors.
\emph{One-stage} detectors attempt to directly classify \textit{and} regress bounding boxes by, either densely classifying a set of predefined anchor boxes~\cite{Redmon16,Redmon2017,Redmon2018,Liu16,Lin2017a,Tan2020} or densely searching for geometric entities of objects \eg~corners, centres or regions~\cite{Law2018,Zhou2019,Tian2019}.
Conversely, most \emph{two-stage} detectors propose class-agnostic bounding boxes using a Region Proposal Network~(RPN),
with predefined sizes and aspect ratios,
filtering out many negative (background) locations.
These bounding box proposals are pooled to region-of-interest (RoI) features and
are classified by a multilayer perceptron (MLP) in the second stage of the detector~\cite{Girshick13,Girshick15,Ren16,Li2019}. \\[-8pt]

\par{\noindent \bf Few-Shot Object Detection}
aims to expand the vocabulary of an object detector by only annotating a handful of samples.
Several works~\cite{Chen2018,Kang2019,Wang2019,Yan2019,Xiao2020,Wang2020,Hu2021,Li2021b,Li2021,Li2021a,Sun2021,Fan2021,Zhu2021,Han2021a,Qiao2021a,Wu2021a,Yang2020,Karlinsky2019,Fan2020}
have been proposed in the recent literature.
The meta-learning method, FSRW~\cite{Kang2019} conditions dense query image features
from a YOLOv2 network~\cite{Redmon2017} with a separate network operating on a support set.
Recently, a simple two-phase fine-tuning approach (TFA) is proposed~\cite{Wang2020},
in which a Faster R-CNN model is initially trained on the base data.
In the second training phase,
only the final classification layer is finetuned on a few samples of novel classes,
with the rest of the model fixed.
This work initiates a shift away from meta-learning based methods for few-shot object detection. FSCE~\cite{Sun2021},
alleviates class confusion between novel classes by training
a separate supervised contrastive learning~\cite{Khosla2020} head on RoI features.
A recent work, DeFRCN~\cite{Qiao2021a},
decouples the training of RPN features and RoI classification.
Another recent work, SRR-FSD~\cite{Zhu2021} combines vision and natural language,
projecting image features into semantic class embeddings learnt from a large text corpus. \\[-8pt]

\par{\noindent \bf Semi-supervised Object Detection}
belongs to another related research area.
Such a problem setup can be traced back to the pre-deep learning era~\cite{Rosenberg05},
where the goal is to train detectors with a combination of labelled, weakly-labelled and unlabelled data.
In the recent literature, the idea of exploiting consistency and self-training has been widely adopted,
for example,
\cite{Jeong19} proposed to enforce the predictions of an input image and its flipped version to be consistent;
\cite{Sohn20, Liu21} pre-trained a detector using a small amount of labelled data and generated pseudo-labels on unlabelled data for further fine-tuning.
Generally speaking,
these methods aim to train a detector on two separate subsets;
one contains images with all objects of all classes being {\em exhaustively} annotated,
and the other subset is fully unlabelled.
Therefore,
the model trained on the labelled set does not suffer the same issue as in FSOD,
where a large number of {\em novel} object instances are wrongly treated as background during base training.
Hence the scarce data issue of FSOD \emph{does} exist in semi-supervised object detection,
but ``supervision collapse'' \emph{does not}. \\[-8pt]

\par{\noindent \bf Self-Training} is a method for gaining noisy pseudo-labels
which has gained renewed interest since it was initially proposed~\cite{Scudder1965}.
In recent years,
self-training has been used to improve image classification by using a teacher-student training regime~\cite{Hinton2015,Chollet2017,Xie2020a}.
This idea is extended to general object detection in~\cite{Zoph2020},
however, their considered scenario is semi-supervised object detection.

% !TEX root = ../main.tex
\def\Dn{$D_{\text{n}}$}
\def\Db{$D_{\text{b}}$}
\def\Dnh{${D}^{*}_{\text{n}}$}
\def\Dnb{$\bar{D}_{\text{n}}$}
\def\Dne{$\tilde{D}_{\text{n}}$}
\def\Dnp{$D^{\prime}_{\text{n}}$}
\def\Bn{$B_{\text{n}}$}
\def\Bb{$B_{\text{b}}$}
\def\Bbb{$\hat{B}_{\text{b}}$}
\def\Bnh{${B}^{*}_{\text{n}}$}
\def\Bnb{$\bar{B}_{\text{n}}$}
\def\Bne{$\tilde{B}_{\text{n}}$}
\def\Bnp{$B^{\prime}_{\text{n}}$}
\def\Snb{$\bar{S}_{\text{n}}$}
\def\Cn{$C_{\text{n}}$}
\def\Cb{$C_{\text{b}}$}

\vspace{-5pt}
\section{Background and Supervision Collapse}
\label{sec:bg}
\vspace{-5pt}

In this section,
we first outline the few-shot object detection task in Section~\ref{ssec:problem_definition}.
Next, in Section~\ref{ssec:improving_tfa},
we carefully examine the various components of the popular two-stage detector~(Faster R-CNN),
identify the critical issues that limit its use in the few-shot scenario and draw conclusions
to ameliorate such critical issues.

\subsection{Problem Definition}
\label{ssec:problem_definition}
In this paper, we consider the same problem setup as in TFA~\cite{Kang2019}.
Specifically,
assuming we are given an image dataset, $\mathcal{D}$,
and two annotation sets.
First, $\mathcal{Y}_{\textsc{base}}$,
with exhaustive annotations on a set of base categories, $\mathcal{C}_{\textsc{base}}$.
Second, $\mathcal{Y}_{\textsc{novel}}^{K}$,
with only $K$ annotations on a set of novel categories, $\mathcal{C}_{\textsc{novel}}$.
Note, the annotations on base categories are exhaustive, but for novel categories
most instances are {\em unlabelled} as only $K$ annotations are provided for the image dataset, $\mathcal{D}$,
under the few-shot setting.

\subsection{Training Strategy}
\label{ssec:improving_tfa}
In this section,
we start by describing a baseline two-stage detector for the problem of few-shot object detection,
following that of TFA~\cite{Wang2020}.
In general, a Faster R-CNN detector, $\Phi_{\textsc{det}}(\cdot)$, can be formulated as:

\vspace{-10pt}
\begin{equation*}
\Phi_{\textsc{det}}(\cdot)= {\Phi_{\textsc{cls}} \circ \Phi_{\textsc{roi}} \circ \Phi_{\textsc{rpn}} \circ \Phi_{\textsc{enc}}(\cdot)}
\end{equation*}
\vspace{-14pt}

\noindent where, each input image is sequentially processed by a set of operations:
an image encoder,~($\Phi_{\textsc{enc}}$); a Region Proposal Network,~($\Phi_{\textsc{rpn}}$);
a region of interest feature module,~($\Phi_{\textsc{roi}}$); and a classification layer on the RoI features,~($\Phi_{\textsc{cls}}$),
mapping to a set of bounding boxes and classes.
Note that, each module here contains the same number of convolutional or MLP layers as the standard Faster R-CNN~\cite{Ren16}.

Training few-shot object detectors involves a two-phase training procedure, as detailed below:  \\[-8pt]

\par{\noindent \bf Base Training:} refers to the standard training for a
Faster R-CNN model~\cite{Ren16,Wu2019a} using {\em base class annotations} only, $\mathcal{Y}_{\textsc{base}}$.
In this work, we do not modify this training regime. \\[-8pt]

\par{\noindent \bf Novel Training:} requires extending the base detector,
such that it can additionally detect instances from novel categories, \ie~$\mathcal{C}_{\textsc{base}} \cup \mathcal{C}_{\textsc{novel}}$.
In recent works, this is usually done by only training (relatively few) layers on novel and base class data,
$\mathcal{Y}_{\textsc{base}}^{K} \cup \mathcal{Y}_{\textsc{novel}}$; the detector is not trained end-to-end on novel class data. For example, in TFA~\cite{Wang2020},
the fewest possible number of parameters are trained on novel class data, namely, $\Phi_{\textsc{cls}}$ only.

Such a two-phase training strategy naturally leads to two questions:
(i) Does an RPN trained on base categories actually generalise, \ie~
are regions proposed for instances of the novel categories?
(ii) How well do features trained only on base categories actually generalise,
in other words, will the RoI features be discriminative for classifying {\em novel} categories?
We aim to answer these two questions on the MS-COCO 30-shot object detection benchmark
(these benchmarks are detailed in Section~\ref{ssec:benchmarks_datasets}).
Specifically, we follow the same data split as in TFA~\cite{Wang2020},
with 60 categories being treated as base categories, and 20 as novel categories.

\vspace{-10pt}
\subsubsection{On Generalisability of RPNs}
\vspace{-5pt}
In standard two-stage object detectors, an RPN is considered as a necessary condition for high-performance detections,
as classification and box coordinate regression will only act on the proposed regions.
Here, we aim to evaluate the quality of an RPN for FSOD, based on recall with respect to the novel categories.

\begin{table}[t]
    \centering
    \scriptsize
    % !TEX root = ../../main.tex
% \begin{table*}[]
% \centering
% \vspace{-28.0pt}
\begin{adjustbox}{width=1.0\linewidth}
\setlength\tabcolsep{4pt}
\begin{tabular}{ccc|cc@{}}
\toprule
 Novel Training?        & \# Proposals  & nAR\textsubscript{IoU=0.5} & $\min$(R50\textsubscript{IoU=0.5}) \\ \midrule
\xmark                  & 100           & 49.7                       & 16.2                               \\
\checkmark              & 100           & 71.0                       & 40.0                               \\ \midrule
Ideal RPN~\cite{Ren16}  & 100           & 84.0                       & 64.5                               \\ \midrule
\xmark                  & 1000          & 82.1                       & 55.9                               \\
\checkmark              & 1000          & 88.5                       & 77.0                               \\ \midrule
Ideal RPN~\cite{Ren16}  & 1000          & 95.0                       & 79.2                               \\  \midrule
\end{tabular}
\end{adjustbox}
\vspace{-10pt}
\caption{
RPN recall evaluation on 20 novel categories from the MS-COCO few-shot object detection benchmark,
using an IoU=$0.5$ criterion.
nAR50 -- novel class average recall, min(nR50) -- minimum novel class recall.
Please refer to the text for detailed discussion.\vspace{-.5cm}}
\label{tab:rpn_finetune}
% \end{table*}

\end{table}

Specifically, we consider the following three settings:
{\em first},
to understand whether an RPN trained on 60 base categories can directly propose novel object instances,
we evaluate the recall of the RPN from the base detector;
{\em second},
% we finetune the RPN~($\Phi_{\textsc{RPN}}$ is composed of 2 convolutional layers)
we finetune the RPN~(composed of 2 convolutional layers)
on both base and given novel categories, \ie~$\mathcal{Y}_{\textsc{base}} \cup \mathcal{Y}_{\textsc{novel}}^{30}$;
{\em third},
% the Ideal RPN which is inherited from an off-the-shelf Faster R-CNN trained on all data for all 80 categories, $\mathcal{Y}_{\textsc{all}}$. \\[-8pt]
the Ideal RPN which is inherited from an off-the-shelf Faster R-CNN trained on exhaustive data for all categories. \\[-8pt]

\par {\noindent \bf Discussion:}
Table~\ref{tab:rpn_finetune} presents the large performance gap between an RPN from the base detector,
and the Ideal RPN with respect to recalling instances from novel categories.
However, finetuning the RPN specific parameters, $\Phi_{\textsc{rpn}}$, on a handful of instances~($K=30$),
yields a substantial increase in average recall~($49.7$ vs.~$71.0$, $82.1$ vs.~$88.5$, for $100$ and $1000$ proposals respectively),
largely bridging the performance gap to the Ideal RPN.
The minimum class recall also increases substantially~($16.2$ vs.~$40.0$, $55.9$ vs.~$77.0$, for $100$ and $1000$ proposals respectively).

\vspace{-10pt}
\subsubsection{On Transferability of Base Features}
\vspace{-5pt}
In this section,
we aim to measure the transferability of the visual features trained on base categories.
Specifically, we keep the encoder~($\Phi_{\textsc{enc}}$) fixed,
and finetune, individual or combinations of, subsequent modules~($\Phi_{\textsc{cls}}, \Phi_{\textsc{roi}}, \Phi_{\textsc{rpn}}$) during {\bf Novel Training}.

\vspace{-2pt}
\begin{table}[!htb]
    \centering
    \scriptsize
    % !TEX root = ../../main.tex
\begin{adjustbox}{width=\linewidth}
\begin{tabular}[t]{c|ccc|ccc}
\toprule
\setlength\tabcolsep{0pt}
& \multicolumn{3}{c|}{Novel Training} & \multicolumn{3}{c}{Metrics}  \\ \toprule
Setting  & $\Phi_{\textsc{cls}}$ & $\Phi_{\textsc{roi}}$ & $\Phi_{\textsc{rpn}}$ & nAP  & nAP50 & nAP75  \\ \midrule
A1       & \checkmark            &  \xmark               &  \xmark               & 13.0 & 24.7  & 12.3   \\
A2       & \checkmark            &  \checkmark           &  \xmark               & 13.3 & 25.6  & 12.6   \\
A3       & \checkmark            &  \checkmark           &  \checkmark           & 14.3 & 27.5  & 12.9   \\ \midrule
A4~(ALL) & \checkmark            &  \xmark               &  \xmark               & 18.3 & 33.3  & 17.7   \\ \midrule
\multicolumn{4}{l|}{Ideal Faster R-CNN~\cite{Ren16}}                             & 43.5 & 67.4  & 46.6   \\ \bottomrule
\end{tabular}
\end{adjustbox}
\vspace{-5pt}
\caption{
Evaluation on the transferability of base features to 20 novel categories from the MS-COCO few-shot object detection benchmark.
During {\bf Novel Training},
we jointly finetune the different modules with the base and novel category data~(only the given few-shot annotations, $\mathcal{Y}_{\textsc{novel}}^{K}$).
As an Oracle test,
we also consider to finetune the classifier with \emph{all} MS-COCO annotations,  \eg~A4~(ALL).
nAP -- novel class average precision.
Please refer to the text for discussion.\vspace{-.1cm}}
\label{tab:tfa_limit}

\end{table}

\vspace{-3pt}
\par {\noindent \bf Discussion:}
As shown in Table~\ref{tab:tfa_limit}, we compare TFA~\cite{Wang2020} (Setting A1)
with finetuning more layers (A3), which tends to be beneficial~($13.0$ vs.~$14.3$ nAP),
however, it remains substantially lower than the Ideal Faster R-CNN~($14.3$ vs.~$43.5$ nAP).
To remove the factor caused by insufficient data annotation,
we compare TFA (A1) with an Oracle test~(A4-ALL) that finetunes the final classifier with exhaustive data on all categories.
However, the result still largely underperforms the Ideal Faster R-CNN reference~($18.3$ vs. $43.5$nAP),
indicating that the feature encoder is heavily biased towards base classes,
and hardly contains discriminative information for classifying instances from novel classes.
This may be expected, as these categories were treated as background during {\bf Base Training}.

These experiments demonstrate the ``supervision collapse'' issue present in FSOD detectors.
We note that ``supervision collapse'' manifests in two ways:
{\em first}, many false positives occur due to class confusion and poor bounding box regression~(poor detection precision),
{\em second}, there are many false negatives or missing detections, despite an improved RPN~(poor detection recall).
We further analyse these manifestations in Appendix~\ref{app:supervision_collapse}.

\vspace{-10pt}
\subsubsection{Summary}
\vspace{-5pt}
After careful evaluations, we draw the following two conclusions:
{\em first},
updating all parameters specific to the RPN~($\Phi_{\textsc{rpn}}$) is essential to improve recall on novel categories;
{\em second}, features trained on base category data are {\bf not} discriminative enough to classify novel instances,
leading to severe performance degradation, and so we update all parameters in the RoI feature module~($\Phi_{\textsc{roi}}$),
in addition to the classification layer~($\Phi_{\textsc{cls}}$).
These two choices constitute our {\bf Novel Training} process, yielding a stronger baseline detector.
This allows enough alleviation of ``supervision collapse'',
such that our baseline detector can be used as a starting point in our pseudo-labelling method,
as will be detailed in the next section.

% !TEX root = ../main.tex

\def\Dn{$D_{\text{n}}$}
\def\Db{$D_{\text{b}}$}
\def\Dnh{${D}^{*}_{\text{n}}$}
\def\Dnb{$\bar{D}_{\text{n}}$}
\def\Dne{$\tilde{D}_{\text{n}}$}
\def\Dnp{$D^{\prime}_{\text{n}}$}
\def\Bn{$B_{\text{n}}$}
\def\Bb{$B_{\text{b}}$}
\def\Bbb{$\hat{B}_{\text{b}}$}
\def\Bnh{${B}^{*}_{\text{n}}$}
\def\Bnb{$\bar{B}_{\text{n}}$}
\def\Bne{$\tilde{B}_{\text{n}}$}
\def\Bnp{$B^{\prime}_{\text{n}}$}
\def\Snb{$\bar{S}_{\text{n}}$}
\def\Cn{$C_{\text{n}}$}
\def\Cb{$C_{\text{b}}$}

\vspace{-5pt}
\section{Method}
\label{sec:method}
\vspace{-5pt}
To address the ``supervision collapse'' issue, we adopt a simple \emph{pseudo-labelling} method for mining instances of novel categories, effectively expanding their annotation set.
However, pseudo-annotations that are na\"ively sourced from the detector~(after {\bf Novel Training}),
are unreliable, containing a large number of false positives.
Here, we establish a method for improving the precision of these candidate pseudo-annotations
by automatically filtering out candidates with incorrect class labels,
and refining the bounding box coordinates for those remaining.
Our method yields a large set of high precision pseudo-annotations for novel categories,
allowing the final detector to be trained end-to-end on both base and novel category data.
We detail the proposed method in the following sections.

% \vspace{-5pt}
\subsection{Candidate Sourcing}
\label{ssec:candidate_sourcing}
\vspace{-3pt}
The goal here is to generate a set of candidate detections that are {\em potentially} valid pseudo-annotations for novel categories.
Specifically, the detector from {\bf Novel Training}~(as described in Section~\ref{ssec:improving_tfa}),
is used to perform inference on the training images~($\mathcal{D}$) to generate a set of candidate detections,
each containing a class label and predicted bounding box coordinates (see Figure~\ref{fig:pipeline}, left).
We limit the size of this set to be 1000s by taking \emph{novel class} detections
with high confidence scores, here we use $q > 0.8$, producing $\hat{\mathcal{Y}}_{\textsc{novel}}$.

As demonstrated in earlier evaluations,
the detector from {\bf Novel Training} cannot detect instances from novel categories well,
leaving a large number of incorrect predictions in the set of candidates,
either misclassification, or imprecise bounding box coordinates.
The key question now becomes, {\em how can we improve the precision of this list?}

% \vspace{-5pt}
\subsection{Label Verification}
\label{ssec:label_verification}
\vspace{-5pt}
In this section we take inspiration from the work on query expansion by Chum~\etal~\cite{Chum07b},
which uses spatial verification to accept or reject new instances during retrieval.
The goal here is to verify the predicted class label for each candidate detection.
Specifically, we consider to build a classifier for the novel categories with the very limited few-shot annotations~($\mathcal{Y}_{\textsc{novel}}$).

Building classifiers with only a few annotations is clearly not a trivial task,
as it often demands high-quality feature representations.
Here, we benefit from the recent development of self-supervised models,
\eg~MoCo~\cite{He2020}, SwAV~\cite{Caron2020}, DINO~\cite{Caron2021a},
and construct $k$NN classifiers with the high-quality features produced from those models.
In practice, this work uses the output \texttt{CLS} token from a ViT model~\cite{Dosovitskiy2020} trained with the self-supervised DINO~\cite{Caron2021a} method,
where the NN performance is shown to be particularly strong.

To perform Label Verification (see Figure~\ref{fig:pipeline}, centre), we first compute features for each of the given novel class groundtruth annotations,
using the self-supervised model.
These features are used as the training data in our $k$NN classifier.
Similarly, we compute features for each instance in the set of candidate detections using the same self-supervised model.
In detail, to compute the feature of a given annotation/candidate detection, we first use the bounding box to crop the relevant image.
This crop is then resized and passed as input to the self-supervised model.

We adopt a simple verification policy:
a given candidate detection is accepted (or verified) if our $k$NN classifier, using cosine similarity,
predicts the same class as the predicted class label from the detector.
With such a verification step, we obtain a {\em verified} set of candidate detections with high precision with respect to classification labels.

\subsection{Box Correction}
\label{ssec:box_correction}
\vspace{-2pt}
In addition to verifying classification labels,
we consider refining the bounding boxes for all remaining candidate detections in the {\em verified} set (see Figure~\ref{fig:pipeline}, right).
Taking inspiration from the Cascade R-CNN~\cite{Cai2018a},
we build a separate model, containing three {\em class-agnostic} regressors that gradually produce a higher-quality bounding box,
with each only processing boxes of similar IOUs to the groundtruth.

Specifically, during {\bf Novel Training},
we divide the RPN proposals into three splits using different IoU thresholds,
and pass the RoI features through the corresponding regressors.
For example, all features pooled from IoU$>0.3$ boxes are passed to the first regressor,
features pooled from IoU$>0.5$ boxes are passed to the second regressor
and features pooled from IoU$>0.7$ boxes are passed to the third regressor.
Once this is trained, the bounding boxes for the {\em verified} set can be corrected,
by feeding their RoI features through the three regressors in succession.

We now possess a large set of {\em previously unlabelled} novel instances,
with high precision class labels and high-quality bounding boxes.
This verified and corrected set is then used as pseudo-annotations to re-train our detector \emph{end-to-end}
on novel and base class instances.

% !TEX root = ../main.tex
\vspace{-5pt}
\section{Experiments}
\label{sec:experiments}
\vspace{-5pt}
In this section,
we first introduce the standard experimental benchmarks used in the literature~\cite{Kang2019,Sun2021}.
After this, in Section~\ref{ssec:implementation} we describe our implementation, training details,
and conduct extensive ablation studies on the design choices of this work in Section~\ref{ssec:ablation}.
Lastly, inheriting the best experience from the ablation studies,
we compare to the existing state-of-the-art approaches in Section~\ref{ssec:fsod_results}.

\vspace{-2pt}
\subsection{Few-Shot Object Detection Benchmarks}
\label{ssec:benchmarks_datasets}
\vspace{-5pt}
We follow the same benchmarks as in~\cite{Kang2019},
evaluating our model on the MS-COCO~\cite{Lin14} and PASCAL VOC~\cite{Everingham15} datasets.
To maintain a fair comparison, we use the same fixed lists of novel samples and data splits given in~\cite{Kang2019}. \\[-8pt]

\par{\noindent \bf MS-COCO}
has $80$ categories in total.
In FSOD, the $20$ categories present in PASCAL VOC are used as novel classes
and the remaining categories are used as base classes.
In this case, the benchmarks are designed for testing with $K=10,30$ shots,
and we report standard MS-COCO metrics, namely Average Precision (IoU=$0.5:0.95$),
Average Precision (IoU=$0.5$) and Average Precision (IoU=$0.75$) on novel classes,
abbreviated to nAP, nAP50 and nAP75, respectively.
Our ablation studies are all conducted on the MS-COCO benchmark. \\[-8pt]

\par{\noindent \bf PASCAL VOC}
contains $20$ classes,
in FSOD, the data is randomly split into $15$ base classes and $5$ novel classes.
There are three such splits and for each novel class there are $K=1,2,3,5,10$ shots available.
For this dataset we report the standard PASCAL VOC metric Average Precision (IoU=$0.5$) for novel classes (nAP50).

\vspace{-2pt}
\subsection{Implementation Details}
\label{ssec:implementation}
\vspace{-1pt}
Experiments are conducted with a standard Faster R-CNN~\cite{Ren16}, with a FPN~\cite{Lin2017}.
All experiments are run on $4$ GPUs with batch-size $16$.
We use a SGD optimiser with momentum $0.9$ and weight decay $10^{-4}$,
except for models with Transformer based backbones,
in which we use the AdamW optimiser with standard hyperparameters and weight decay $0.05$, following~\cite{Liu2021}.
The number of finetuning iterations is scaled depending on the dataset and the number of available shots.

We apply \texttt{RandomCrop} and \texttt{ColorJitter} augmentations when finetuning our FSOD detector on novel class data, unless stated otherwise.
For experiments on MS-COCO,
we include the \texttt{Mosaic} augmentation introduced in YOLOv4~\cite{Bochkovskiy2020}.
This augmentation helps improve detection performance on ``small'' objects by stitching $4$ images into a
$2\times2$ grid.
\texttt{Mosaic} is not used on PASCAL VOC experiments as the dataset does not contain the same scale variation as MS-COCO.

The number of neighbours $k$ used in Label Verification (Section~\ref{ssec:label_verification}) is determined by the number of novel instances, $k$$=$$\min\left(\left\lfloor\frac{K}{3}\right\rfloor + 1, 10\right)$.
Hence for $K$~$=$~$1,2,3,5,10,30$, $k$~$=$~$1,1,2,2,4,10$.
We choose a self-supervised DINO ViT-S/8~\cite{Dosovitskiy2020, Caron2021a} to construct the kNN classifier for verification,
using the output \texttt{CLS} token as the chosen feature.
For Box Correction (Section~\ref{ssec:box_correction}), we train a series of three box regression heads,
where positive boxes are defined as those with $\text{IoU} >(0.3,0.5,0.7)$, respectively.
This choice enables the correction of relatively poor initial bounding boxes.
% \weidi{a bit redundant to methods, but I think it's OK.}

After Candidate Sourcing, Label Verification and Box Correction,
we significantly increase the number of available samples for novel categories,
however it is inevitable that many novel instances will remain absent in our pseudo-annotations,
and would still be treated as background.
To avoid this issue, we introduce {\em ``ignore regions''}, during end-to-end re-training,
which are considered as neither foreground nor background,
in practice, we treat all {\em unverified} novel class detections as ignore regions.
We analyse these {\em ``ignore regions''} in Appendix~\ref{app:ignore_regions}.

% -----------------------------------------
% !TEX root = ../../main.tex
\begin{table*}[!htb]
% \vspace{8pt}
\begin{minipage}[t]{0.44\linewidth}
\scriptsize
\setlength\tabcolsep{1.5pt}
\centering
\begin{adjustbox}{width=\columnwidth}
\begin{tabular}[t]{@{}c|ccc|c|c|c|c|@{}}
\toprule
\multirow{2}{*}{Setting} & \multicolumn{3}{c|}{Augmentations}                                                                                                                                     & \multirow{2}{*}{\begin{tabular}[c]{@{}c@{}}Finetune\\ $\Phi_{\textsc{rpn}}$, $\Phi_{\textsc{roi}}$?\end{tabular}} & \multirow{2}{*}{\begin{tabular}[c]{@{}c@{}}$\Phi_{\textsc{roi}}$\\ Dropout?\end{tabular}} & \multirow{2}{*}{nAP} & \multirow{2}{*}{nAPs} \\
\                        &  \begin{tabular}[c]{@{}c@{}}\texttt{Color}\\ \texttt{Jitter} \end{tabular} & \begin{tabular}[c]{@{}c@{}}\texttt{Random}\\ \texttt{Crop} \end{tabular} & \texttt{Mosaic} &                                                                                                                   &                                                                                           &                      &                       \\ \midrule
B1                       &                                                                           &                                                                          &                 &                                                                                                                   &                                                                                           & 13.0                 & 5.5                   \\
B2                       & \checkmark                                                                &                                                                          &                 &                                                                                                                   &                                                                                           & 13.2                 & 5.6                   \\
B3                       &                                                                           & \checkmark                                                               &                 &                                                                                                                   &                                                                                           & 13.2                 & 5.6                   \\
B4                       &                                                                           &                                                                          & \checkmark      &                                                                                                                   &                                                                                           & 12.8                 & 6.1                   \\
B5                       & \checkmark                                                                & \checkmark                                                               & \checkmark      &                                                                                                                   &                                                                                           & 13.8                 & 6.8                   \\ \midrule
C1                       & \checkmark                                                                & \checkmark                                                               &                 & \checkmark                                                                                                        &                                                                                           & 14.5                 & 4.3                   \\
C2                       & \checkmark                                                                & \checkmark                                                               & \checkmark      & \checkmark                                                                                                        &                                                                                           & 16.2                 & 6.2                   \\
C3                       & \checkmark                                                                & \checkmark                                                               & \checkmark      & \checkmark                                                                                                        & \checkmark                                                                                  & 16.6                 & 7.0                   \\ \bottomrule
\end{tabular}
\end{adjustbox}
\vspace{-7pt}
\captionof{table}{\small{Few-shot object detection performance on \textit{novel} classes on the MS-COCO 30-shot task.
Note that, the baseline model B1 is equivalent to a strong baseline proposed in TFA~\cite{Wang2020}.
nAP -- novel class average precision (nAPs is for small instances only)
}}
\label{tab:ablation_mid}

\vspace{-1.5pt}
\begin{adjustbox}{width=.98\columnwidth}
\begin{tabular}[t]{@{}l|ccc|ccc@{}}
\toprule
Method                           & nAP  & nAP50 & nAP75 & bAP  & bAP50 & bAP75 \\ \midrule
Baseline                         & 16.6 & 30.9  & 15.8  & 29.5 & 46.7  & 32.4  \\
+ Candidate Sourcing             & 18.4 & 34.0  & 18.2  & 31.7 & 53.2  & 35.6  \\
+ Label Verification             & 21.8 & 40.3  & 21.1  & 31.6 & 53.4  & 35.2  \\
+ Box Correction                 & 25.5 & 42.0  & 27.3  & 33.3 & 53.6  & 36.1  \\ \midrule
Ideal Faster R-CNN~\cite{Ren16}  & 43.5 & 67.4  & 46.6  & 36.1 & 56.0  & 39.2  \\ \bottomrule
\end{tabular}
\end{adjustbox}
\vspace{-8pt}
\captionof{table}{\small{Ablation study on our pseudo-labelling method.}\vspace{-.2cm}}
\label{tab:ablation_qe_steps}
\end{minipage}
% ==================================================
\hfill
\begin{minipage}[t]{.54\linewidth}
\centering
\scriptsize
\begin{adjustbox}{width=\columnwidth}
\setlength\tabcolsep{2.5pt} % default value: 6pt
\begin{tabular}[t]{@{}l|c|cc|cc|cc|cc@{}}
\toprule
\multirow{2}{*}{Method}          & \multirow{2}{*}{Backbone}     & \multicolumn{2}{c|}{nAP}      & \multicolumn{2}{c|}{nAP50}      & \multicolumn{2}{c}{nAP75}      & \multicolumn{2}{c}{bAP}       \\
                                 &                               & 10            & 30            & 10             & 30             & 10             & 30            & 10           & 30             \\ \midrule
% FSRW~\cite{Kang2019}             & DarkNet-19                   & 5.6           & 9.1           & 12.3           & 19.0           & 4.6            & 7.6           & -            & -              \\ \midrule
CGDP+FSCN~\cite{Li2021}          &                               & 11.3          & 15.1          & 20.3           & 29.4           & -              & -             & -            & -              \\
\rowcolor[HTML]{EFEFEF}
Ours (Baseline)                  & ResNet-50                     & \sbest{11.4}  & \sbest{16.6}  & \sbest{21.2}   & \sbest{30.9}   & \sbest{11.1}   & \sbest{15.8}  & \sbest{28.4} & \sbest{29.5}   \\
\rowcolor[HTML]{EFEFEF}
Ours (Pseudo-Labelling)          &                               & \best{17.6}   & \best{25.5}   & \best{30.9}    & \best{42.0}    & \best{17.3}    & \best{27.3}   & \best{29.7}  & \best{33.3}    \\ \midrule
% MetaDet~\cite{Wang2019}          & \multirow{14}{*}{ResNet-101} & 7.1           & 11.3          & 14.6           & 21.7           & 6.1            & 8.1           & -            & -              \\
% Meta R-CNN~\cite{Yan2019}        &                              & 8.7           & 12.4          & 19.1           & 25.3           & 6.6            & 10.8          & -            & -              \\
% MPSR~\cite{Wu2020a}              &                              & 9.8           & 14.1          & 17.9           & 25.4           & 9.7            & 14.2          & -            & -              \\
% TFA w/ fc~\cite{Wang2020}        &                              & 10.0          & 13.4          & 19.2           & 24.7           & 9.2            & 13.2          & 32.0         & 33.8           \\
TFA w/ cos~\cite{Wang2020}       &                               & 10.0          & 13.7          & 19.1           & 24.9           & 9.3            & 13.4          & \sbest{32.4} & \sbest{34.2}   \\
FCSE~\cite{Sun2021}              &                               & 11.9          & 16.4          & -              & -              & 10.5           & 16.2          & -            & -              \\
Retentive R-CNN~\cite{Fan2021}   &                               & 10.5          & 13.8          & -              & -              & -              & -             & \best{39.2}  & \best{39.3}    \\
SRR-FSD~\cite{Zhu2021}           &                               & 11.3          & 14.7          & 23.0           & 29.2           & 9.8            & 13.5          & -            & -              \\
CME~\cite{Li2021b}               &                               & 15.1          & 16.9          & 24.6           & 28.0           & \sbest{16.4}   & \sbest{17.8}  & -            & -              \\
DCNet~\cite{Hu2021}              &                               & 12.8          & 18.6          & 23.4           & 32.6           & 11.2           & 17.5          & -            & -              \\
TIP~\cite{Li2021a}               &                               & 16.3          & 18.3          & \best{33.2}    & \sbest{35.9}   & 14.1           & 16.9          & -            & -              \\
QA-FewDet~\cite{Han2021a}        &                               & 11.6          & 16.5          & 23.9           & 31.9           & 9.8            & 15.5          & -            & -              \\
FSOD-UP~\cite{Wu2021a}           &                               & 11.0          & 15.6          & -              & -              & 10.7           & 15.7          & -            & -              \\
DeFRCN    ~\cite{Qiao2021a}      &                               & \best{18.5}   & \sbest{22.6}  & -              & -              & -              & -             & N/A          & N/A            \\
\rowcolor[HTML]{EFEFEF}
Ours (Baseline)                  &                               & 12.1          & 17.8          & 22.0           & 31.6           & 11.8           & 17.7          & 31.9         & 31.8           \\
\rowcolor[HTML]{EFEFEF}
Ours (Pseudo-Labelling)          & \multirow{-12}{*}{ResNet-101} & \sbest{17.8}  & \best{24.5}   & \sbest{30.9}   & \best{41.1}    & \best{17.8}    & \best{25.0}   & 31.9         & 33.0           \\ \midrule
\rowcolor[HTML]{C0C0C0}
Ours (Baseline)                  &                               & 12.6          & 19.0          & 23.3           & 35.3           & 12.1           & 18.3          & 27.2         & 31.5           \\
\rowcolor[HTML]{C0C0C0}
Ours (Pseudo-Labelling)          & \multirow{-2}{*}{Swin-T}      & 18.6          & 26.1          & 32.1           & 45.0           & 18.5           & 26.8          & 29.2         & 31.1           \\ \midrule
\rowcolor[HTML]{C0C0C0}
Ours (Baseline)                  &                               & 14.4          & 20.3          & 25.8           & 37.2           & 14.4           & 20.2          & 33.0         & 36.2           \\
\rowcolor[HTML]{C0C0C0}
Ours (Pseudo-Labelling)          & \multirow{-2}{*}{Swin-S}      & 19.0          & 26.8          & 34.1           & 45.8           & 19.0           & 27.5          & 28.7         & 34.8           \\ \bottomrule
% Ours (Query Expand 2x)  & \textit{}         & tbc           & 16.7          & tbc            & 31.1           & tbc            & 16.2          \\ \bottomrule
\end{tabular}
\end{adjustbox}
\vspace{-8pt}
\captionof{table}{\small{Few-shot detection performance on the MS-COCO benchmark.
We report performance on the 20 novel classes of MS-COCO in the FSOD setting.
Best and second-best results,  per backbone, are coloured \best{blue} and \sbest{red}, respectively.
We also report performance on base classes.}}%\vspace{-.2cm}}
% \vspace{-15pt}
\label{tab:coco_novel_post}
\end{minipage}
\vspace{-12pt}
\end{table*}
% ==================================================

% -----------------------------------------

\vspace{-2pt}
\subsection{Ablation Studies}
\label{ssec:ablation}
We conduct ablation studies to investigate our design choices.
The following experiments are considered:
\emph{first}, we demonstrate the importance of data augmentation to yield a stronger baseline model before any of the pseudo-labelling steps;
\emph{second}, we analyse several critical components of our method and conduct thorough ablation studies to validate their necessity, namely, Candidate Sourcing, Label Verification, Box Correction;
\emph{third}, we show that our proposed approach maintains performance for base class detections.
Note that, all ablation experiments are conducted on the MS-COCO benchmark with $K=30$ and a ResNet-50 backbone. \\[-10pt]

\par{\noindent \bf Importance of Augmentations:}
Given we only have access to a limited number of samples for novel categories at the starting point,
maximising data efficiency before any pseudo-labelling is critical.
Table~\ref{tab:ablation_mid} presents our observations.
When comparing to TFA~\cite{Wang2020} as a baseline model~(equivalent to Setting B1),
applying \texttt{ColorJitter}, \texttt{RandomCrop} and \texttt{Mosaic} augmentations~(Settings B2-B4)
yields negligible performance improvements.
Since almost all layers in TFA have been frozen during {\bf Novel Training},
augmentations can only affect the classification layer of Faster R-CNN, $\Phi_{\textsc{cls}}$.
The combination of these three augmentations (Setting B5), only gives a marginal improvement of $0.8$nAP.

As explained in Section~\ref{ssec:improving_tfa},
we improve the {\bf Novel Training} stage by also updating all RoI parameters and the RPN,
{\em i.e.}~$\Phi_{\textsc{RPN}}, \Phi_{\textsc{ROI}}$.
With this simple change, all augmentations (Setting C2) substantially improve results,
yielding a $3.2$nAP boost over the TFA baseline model (Setting B1).
In addition, we observe a noticeable improvement in performance on small novel instances ($6.2$ vs.~$4.3$nAPs)
from \texttt{Mosaic} augmentations, {\em e.g.}~B5 vs.~B1-B3, and C2 vs.~C1.
To further combat the overfitting issue, we add Dropout on RoI activations (Setting C3),
yielding a small additional improvement of $0.4$nAP.\\[-10pt]

\par{\noindent \bf Ablation of Pseudo-Labelling Steps:}
Table~\ref{tab:ablation_qe_steps} shows the importance of our overall pseudo-labelling method and the contribution of each step.
Using the Candidate Sourcing step only is equivalent to treating the na\"ive detections as pseudo-annotations,
% \weidi{this is very confusing now, as our method is called ``pseudo-labelling'' as well.}
as done in other self-training works~\cite{Liu21,Sohn20};
this gives a marginal performance improvement of $1.8$nAP.
Removing the pseudo-annotations with incorrect class labels by Label Verification,
yields an additional $3.4$nAP performance boost.
Lastly, Box Correction brings a $3.7$nAP performance boost,
in particular, such improvement is largely attributed to that from the stricter metric, {\em i.e.}~nAP75,
clearly showing the reduction of bounding box regression errors.
Note that,
the number and distribution of class labels for pseudo-annotations are identical before and after Box Correction;
only the box coordinates of each pseudo-annotation have been changed. \\[-10pt]

\noindent {\bf Effect on Base Class Performance:}
While reading the performance on base classes from Table~\ref{tab:ablation_qe_steps},
we observe that, for the \texttt{Baseline} model,
the improved performance on novel classes comes at the cost of performance on base classes,
{\em e.g.}, bAP drops from 36.1 to 29.5.
Our proposed pseudo-labelling method improves the detection of novel classes, while recovering performance on base classes.

\vspace{-5pt}
\subsection{Comparison to SotA}
\label{ssec:fsod_results}
\vspace{-5pt}
\par{\noindent \bf Existing methods}
include the meta-learning approaches:
\texttt{CGDP+FSCN}~\cite{Li2021}, \texttt{CME}~\cite{Li2021b}, \texttt{TIP}~\cite{Li2021a}, \texttt{DCNet}~\cite{Hu2021},
and two-phase training works:~\texttt{TFA}~\cite{Wang2020}, \texttt{FCSE}~\cite{Sun2021}, \texttt{Retentive R-CNN}~\cite{Fan2021},
\texttt{SRR-FSD}~\cite{Zhu2021}, \texttt{DeFRCN}~\cite{Qiao2021a}, \texttt{FSOD-UP}~\cite{Wu2021a}, \texttt{QA-FewDet}~\cite{Han2021a}.
In particular, we note that very few works report results on the base class detection performance,
and methods like~\texttt{DeFRCN}~\cite{Qiao2021a},
can actually only detect novel classes and does not maintain the ability to detect base classes as with ours.

We report two sets of results for each task,
\texttt{Baseline} which makes use
of augmentations and the improved training outlined in Section~\ref{sec:bg} and \texttt{Pseudo-Labelling} which follows the method as outlined in Section~\ref{sec:method}. \\[-10pt]

\par{\noindent \bf MS-COCO Results}
are shown in Table~\ref{tab:coco_novel_post}.
% \weidi{say which backbone are you discussing ?}
Using a ResNet-50 backbone, our \texttt{Baseline} method, which makes extensive use of augmentations and improved {\bf Novel Training},
outperforms many existing works for $K=30$, reflecting the importance of our findings in Section~\ref{ssec:improving_tfa}.
When applying our \texttt{Pseudo-Labelling} method,
we set new SotA performance for $K$$=$$30$,
with a performance boost of up to $2.9$, $6.1$ and $10.4$ for nAP, nAP50 and nAP75 metrics, respectively.
When $K$$=$$10$, with a ResNet-101 backbone, our \texttt{Pseudo-Labelling} method achieves state-of-the-art or second-best performance in terms of
nAP, nAP50, nAP75. In terms of nAP, only \texttt{DeFRCN} outperforms our work.
However, note that, \texttt{DeFRCN} is not able to detect base classes after training on novel categories,
which strives for a different purpose than ours, we aim to {\em expand} our detector,
rather than to {\em transfer} the detector.

In addition, while using a more powerful Transformer backbone models (Swin-T, Swin-S~\cite{Liu2021}),
our proposed pseudo-labelling provides additional performance boosts. \\[-8pt]

\par{\noindent \bf PASCAL VOC Results}
are shown for the three standard novel splits in Table~\ref{tab:pascal_new_post}.
Incorporating our proposed \texttt{Pseudo-Labelling} method
is among the top 2 best performing models in all cases (except Novel Split 2 for $K$$=$$2$),
and sets SotA performance for the majority of cases.

Notably, for Novel Split 3, our \texttt{Pseudo-Labelling} method achieves SotA for all $K$,
with a boost of up to $4.7$ nAP50, when using a ResNet-101 backbone.
In many cases our weaker detection backbone (ResNet-50) already achieves SotA performance.
Note that, once again the transferred detector \texttt{DeFRCN} outperforms ours in some cases.

\vspace{.1cm}
\subsection{Qualitative Results}
\begin{figure*}[!htb]
\centering
\scriptsize
\includegraphics[width=\linewidth]{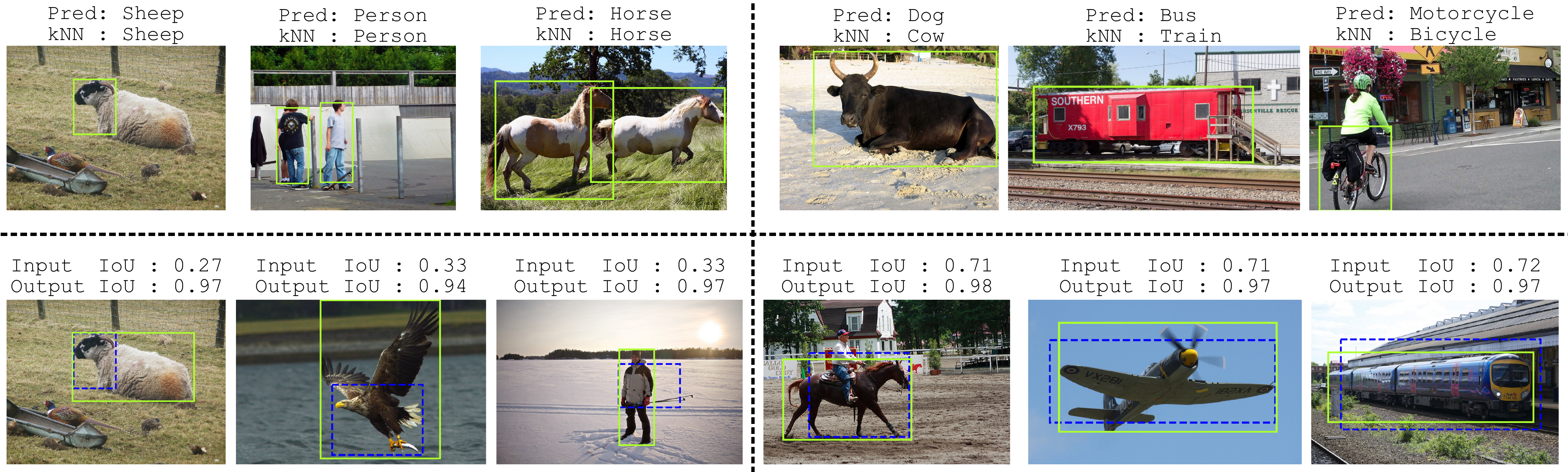}
\vspace{-0.4cm}
\caption{
Top Left: Predicted instances which are verified during {\em Label Verification};
the predicted class labels from our baseline detector and our $k$NN classifier match.
Top Right: Predicted instances which are rejected during {\em Label Verification};
the predicted class labels from our baseline detector (false positive) and the $k$NN {\em do not} match.
Bottom Left: Verified bounding boxes with very poor quality (blue dashed) are drastically improved (lime solid) during {\em Box Correction}.
Bottom Right: Verified bounding boxes which are acceptable (blue dashed) are further improved (lime solid).\vspace{-0.2cm}}
% \caption{\hl{TO DO}}
\label{fig:verify_correct}
\end{figure*}

In Figure~\ref{fig:verify_correct}, we present the qualitative results after each step of our pseudo-labelling procedure.
The top row of Figure~\ref{fig:verify_correct} shows examples of Label Verification (Section~\ref{ssec:label_verification}).
The first three examples demonstrate the case in which the predicted class label from our detector matches $k$NN classification,
and so the candidate detection is verified.
The last three examples show the opposite case in which candidate detections are correctly rejected.
The bottom row of Figure~\ref{fig:verify_correct} shows examples of Box Correction~(Section~\ref{ssec:box_correction}).
The first three examples show very poor bounding boxes from verified candidates (dashed blue boxes),
which are drastically improved during the Box Correction step.
The last three examples show acceptable bounding boxes from verified candidates (dashed blue boxes),
also being improved with Box Correction.
This demonstrates the ability of our Box Correction model to deal with a wide range of bounding box quality with respect to input candidate detections.

\begin{figure*}[!htb]
\centering
\scriptsize
\includegraphics[width=\linewidth]{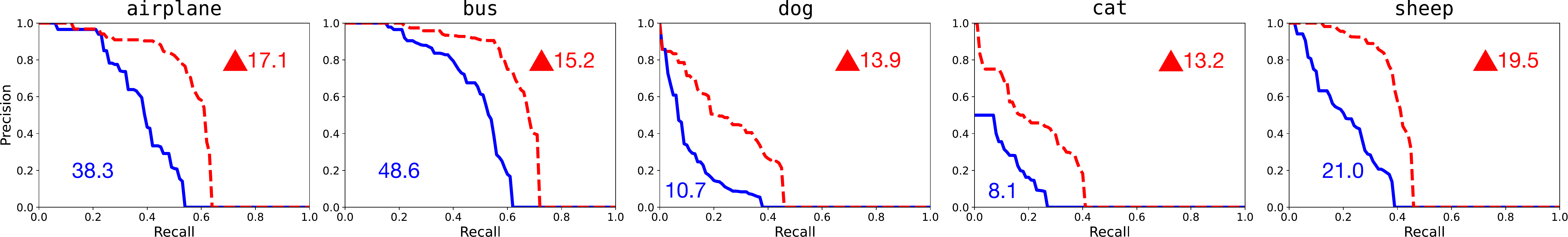}
\vspace{-0.4cm}
\caption{Precision-Recall curves (using the stricter IoU$=$$0.75$ criterion) for $K$$=$$30$ on MS-COCO 
showing Baseline performance (solid {\color[HTML]{3531FF}{blue}})
and the substantial performance boost after making use of our Pseudo-Labelling method (dashed {\color[HTML]{FE0000}{red}}).
Our Pseudo-Labelling method yields improved precision and improved recall for novel classes.\vspace{-.3cm}}
\label{fig:pr_curves}
\end{figure*}

In Figure~\ref{fig:pr_curves}, we show precision-recall curves for some novel classes on the MS-COCO benchmark for $K$$=$$30$ using the stricter IoU$=$$0.75$ criterion. 
Our Pseudo-Labelling method substantially improves novel class performance, with improved precision and novel class recall.
We note that for many novel classes, baseline models suffer from poor recall due to limited novel class annotations.
This poor recall is improved by our pseudo-labelling method, however the poor recall of the baseline detector puts a
limit on the diversity of pseudo-annotations for end-to-end retraining.

% -----------------------------------------
% !TEX root = ../../main.tex
\newcommand\hfilll{\hspace{0pt plus 1filll}}
\begin{table*}[!htb]
% \vspace{10pt}
\centering
\scriptsize
\begin{adjustbox}{width=\textwidth}
\setlength\tabcolsep{3.5pt}
\begin{tabular}{@{}l|c|ccccc|ccccc|ccccc@{}}
\toprule
\multirow{2}{*}{Method/Shot}                             & \multirow{2}{*}{Backbone}    & \multicolumn{5}{c|}{Novel Split 1}                                       & \multicolumn{5}{c|}{Novel Split 2}                                       & \multicolumn{5}{c}{Novel Split 3}                                        \\
                                                         &                              & 1            & 2            & 3            & 5            & 10           & 1            & 2            & 3            & 5            & 10           & 1            & 2            & 3            & 5            & 10           \\ \midrule
% FSRW~\cite{Kang2019}            \hfilll \textit{ICCV 19} & Darknet-19                   & 14.8         & 15.5         & 26.7         & 33.9         & 47.2         & 15.7         & 15.3         & 22.7         & 30.1         & 40.5         & 21.3         & 25.6         & 28.4         & 42.8         & 45.9         \\ \midrule
CGDP+FSCN~\cite{Li2021}         \hfilll \textit{CVPR 21} &                              & 40.7         & 45.1         & 46.5         & 57.4         & 62.4         & 27.3         & 31.4         & 40.8         & 42.7         & 46.3         & 31.2         & 36.4         & 43.7         & 50.1         & 55.6         \\
\rowcolor[HTML]{EFEFEF}
Ours (Baseline)                 \hfilll                  &                              & 37.7         & 42.0         & 50.3         & 57.0         & 58.0         & 19.8         & 22.8         & 35.6         & 42.7         & 44.2         & 33.9         & 36.0         & 38.6         & 49.8         & 51.6         \\
\rowcolor[HTML]{EFEFEF}
Ours (Pseudo-Labelling)         \hfilll                  & \multirow{-3}{*}{ResNet-50}  & 50.5         & 53.1         & 56.4         & 61.7         & 62.7         & \best{36.4}  & 33.8         & 46.1         & 49.3         & 48.2         & \sbest{42.4} & 44.3         & 49.1         & \sbest{55.2} & \sbest{57.6} \\ \midrule
% MetaDet~\cite{Wang2019}         \hfilll \textit{ICCV 19} &                              & 18.9         & 20.6         & 30.2         & 36.8         & 49.6         & 21.8         & 23.1         & 27.8         & 31.7         & 43.0         & 20.6         & 23.9         & 29.4         & 43.9         & 44.1         \\
% Meta R-CNN~\cite{Yan2019}       \hfilll \textit{ICCV 19} &                              & 19.9         & 25.5         & 35.0         & 45.7         & 51.5         & 10.4         & 19.4         & 29.6         & 34.8         & 45.4         & 14.3         & 18.2         & 27.5         & 41.2         & 48.1         \\
% TFA w/ fc~\cite{Wang2020}       \hfilll \textit{ICML 20} &                              & 36.8         & 29.1         & 43.6         & 55.7         & 57.0         & 18.2         & 29.0         & 33.4         & 35.5         & 39.0         & 27.7         & 33.6         & 42.5         & 48.7         & 50.2         \\
TFA w/ cos~\cite{Wang2020}      \hfilll \textit{ICML 20} &                              & 39.8         & 36.1         & 44.7         & 55.7         & 56.0         & 23.5         & 26.9         & 34.1         & 35.1         & 39.1         & 30.8         & 34.8         & 42.8         & 49.5         & 49.8         \\
% MPSR~\cite{Wu2020a}             \hfilll \textit{ECCV 20} &                              & 41.7         & -            & 51.4         & 55.2         & 61.8         & 24.4         & -            & 39.2         & 39.9         & 47.8         & 35.6         & -            & 42.3         & 48.0         & 49.7         \\
FSCE~\cite{Sun2021}             \hfilll \textit{CVPR 21} &                              & 44.2         & 43.8         & 51.4         & 61.9         & \sbest{63.4} & 27.3         & 29.5         & 43.5         & 44.2         & 50.2         & 37.2         & 41.9         & 47.5         & 54.6         & 58.5         \\
Retentive R-CNN~\cite{Fan2021}  \hfilll \textit{CVPR 21} &                              & 42.4         & 45.8         & 45.9         & 53.7         & 56.1         & 21.7         & 27.8         & 35.2         & 37.0         & 40.3         & 30.2         & 37.6         & 43.0         & 49.7         & 50.1         \\
SRR-FSD~\cite{Zhu2021}          \hfilll \textit{CVPR 21} &                              & 47.8         & 50.5         & 51.3         & 55.2         & 56.8         & 32.5         & 35.3         & 39.1         & 40.8         & 43.8         & 40.1         & 41.5         & 44.3         & 46.9         & 46.4         \\
CME~\cite{Li2021b}              \hfilll \textit{CVPR 21} &                              & 41.5         & 47.5         & 50.4         & 58.2         & 60.9         & 27.2         & 30.2         & 41.4         & 42.5         & 46.8         & 34.3         & 39.6         & 45.1         & 48.3         & 51.5         \\
DCNet~\cite{Hu2021}             \hfilll \textit{CVPR 21} &                              & 33.9         & 37.4         & 43.7         & 51.1         & 59.6         & 23.2         & 24.8         & 30.6         & 36.7         & 46.6         & 32.3         & 34.9         & 39.7         & 42.6         & 50.7         \\
TIP~\cite{Li2021a}              \hfilll \textit{CVPR 21} &                              & 27.7         & 36.5         & 43.3         & 50.2         & 59.6         & 22.7         & 30.1         & 33.8         & 40.9         & 46.9         & 21.7         & 30.6         & 38.1         & 44.5         & 50.9         \\
QA-FewDet~\cite{Han2021a}       \hfilll \textit{ICCV 21} &                              & 42.4         & 51.9         & 55.7         & 62.6         & \sbest{63.4} & 25.9         & \sbest{37.8} & 46.6         & 48.9         & \best{51.1}  & 35.2         & 42.9         & 47.8         & 54.8         & 53.5         \\
FSOD-UP~\cite{Wu2021a}          \hfilll \textit{ICCV 21} &                              & 43.8         & 47.8         & 50.3         & 55.4         & 61.7         & 31.2         & 30.5         & 41.2         & 42.2         & 48.3         & 35.5         & 39.7         & 43.9         & 50.6         & 53.5         \\
DeFRCN~\cite{Qiao2021a}         \hfilll \textit{ICCV 21} &                              & \sbest{53.6} & \best{57.5}  & \best{61.5}  & \best{64.1}  & 60.8         & 30.1         & \best{38.1}  & \sbest{47.0} & \best{53.3}  & 47.9         & \best{48.4}  & \sbest{50.9} & \sbest{52.3} & 54.9         & 57.4         \\
\rowcolor[HTML]{EFEFEF}
Ours (Baseline)                 \hfilll                  &                              & 36.0         & 40.1         & 48.6         & 57.0         & 59.9         & 22.3         & 22.8         & 39.2         & 44.2         & 47.8         & 34.3         & 43.4         & 42.9         & 52.0         & 54.5         \\
\rowcolor[HTML]{EFEFEF}
Ours (Pseudo-Labelling)         \hfilll                  & \multirow{-12}{*}{ResNet-101}& \best{54.5}  & \sbest{53.2} & \sbest{58.8} & \sbest{63.2} & \best{65.7}  & \sbest{32.8} & 29.2         & \best{50.7}  & \sbest{49.8} & \sbest{50.6} & \best{48.4}  & \best{52.7}  & \best{55.0}  & \best{59.6}  & \best{59.6}  \\ \bottomrule
\end{tabular}
\end{adjustbox}
\vspace{-0.2cm}
\caption{Few-shot detection performance across the three splits on the PASCAL VOC benchmark.
Best and second-best results are coloured \best{blue} and \sbest{red}, respectively.
Please refer to the text for discussion. \vspace{-0.2cm}}
\label{tab:pascal_new_post}
\end{table*}

\vspace{-0.4cm}

% -----------------------------------------

% !TEX root = ../main.tex
\vspace{0.2cm}
\section{Conclusion}
\label{sec:conclusion}
In this paper, we tackle the problem of few-shot object detection by training on pseudo-annotations.
We present two novel methods to improve the precision of the pseudo-labelling procedure:
{\em first}, we use the given $K$ few-shot annotations to construct classifiers to
verify class labels sourced from a baseline detector;
{\em second}, we train a specialised box correction model to drastically improve
the precision of pseudo-annotation bounding box coordinates.
Our method generates a large number of high-precision pseudo-annotations with precise bounding boxes,
removing the class imbalance issue in FSOD.
This enables re-training of our detector {\em end-to-end},
alleviating the identified issues around ``supervision collapse'' in few-shot object detectors.
Furthermore, we have illustrated the importance of augmentations for FSOD,
this was previously under-explored, despite augmentations being a key part of preventing overfitting.
Our method achieves state-of-the-art or second-best performance performance on both PASCAL VOC and MS-COCO benchmarks,
across all number of shots.

\vspace{-5pt}
\section{Acknowledgements}
\vspace{-6pt}
We thank Rhydian Windsor, Sagar Vaze, Ragav Sachdeva and Guanqi Zhan
for help with proofreading.
This research was supported by the EPSRC CDT in AIMS EP{\small$\slash$}L015897{\small$\slash$}1 and EP{\small$\slash$}S515541{\small$\slash$}1, EPSRC Programme Grant VisualAI EP{\small$\slash$}T028572{\small$\slash$}1,  and a Royal Society Research Professorship RP{\small$\backslash$}R1{\small$\backslash$}191132.
%Weidi Xie would like to acknowledge the generous support of Yimeng Long and Yike Xie in enabling his contribution.

%%%%%%%%% REFERENCES
{
    % \clearpage
    \small
    \bibliographystyle{ieee_fullname}
    \bibliography{longstrings,vgg_local,vgg_other,external}

\begin{thebibliography}{10}\itemsep=-1pt

\bibitem{Bochkovskiy2020}
Alexey Bochkovskiy, Chien-Yao Wang, and Hong-Yuan~Mark Liao.
\newblock Yolov4: Optimal speed and accuracy of object detection.
\newblock {\em arXiv preprint arXiv:2004.10934}, 2020.

\bibitem{Cai2018a}
Zhaowei Cai and Nuno Vasconcelos.
\newblock Cascade r-cnn: Delving into high quality object detection.
\newblock In {\em Proceedings of the IEEE Conference on Computer Vision and
  Pattern Recognition}, pages 6154--6162. IEEE Computer Society, 2018.

\bibitem{Caron2020}
Mathilde Caron, Ishan Misra, Julien Mairal, Priya Goyal, Piotr Bojanowski, and
  Armand Joulin.
\newblock Unsupervised learning of visual features by contrasting cluster
  assignments.
\newblock {\em arXiv preprint arXiv:2006.09882}, 2020.

\bibitem{Caron2021a}
Mathilde Caron, Hugo Touvron, Ishan Misra, Herv\'e J\'egou, Julien Mairal,
  Piotr Bojanowski, and Armand Joulin.
\newblock Emerging properties in self-supervised vision transformers.
\newblock {\em arXiv preprint arXiv:2104.14294}, 2021.

\bibitem{Chen2018}
Hao Chen, Yali Wang, Guoyou Wang, and Yu Qiao.
\newblock Lstd: A low-shot transfer detector for object detection.
\newblock 2018.

\bibitem{Chollet2017}
François Chollet.
\newblock Xception: Deep learning with depthwise separable convolutions.
\newblock In {\em Proceedings of the IEEE Conference on Computer Vision and
  Pattern Recognition}, 2017.

\bibitem{Chum07b}
Ondrej Chum, James Philbin, Josef Sivic, Michael Isard, and Andrew Zisserman.
\newblock Total recall: {A}utomatic query expansion with a generative feature
  model for object retrieval.
\newblock In {\em Proceedings of the 11th International Conference on Computer
  Vision, Rio de Janeiro, Brazil}, 2007.

\bibitem{Dai2021}
Xiyang Dai, Yinpeng Chen, Bin~(Leo) Hsiao, Dongdong Chen, Mengchen Liu, Lu
  Yuan, and Lei Zhang.
\newblock Dynamic head: Unifying object detection heads with attentions.
\newblock In {\em Proceedings of the IEEE Conference on Computer Vision and
  Pattern Recognition}, June 2021.

\bibitem{Doersch2020}
Carl Doersch, Ankush Gupta, and Andrew Zisserman.
\newblock Crosstransformers: spatially-aware few-shot transfer.
\newblock In {\em Advances in Neural Information Processing Systems}, 2020.

\bibitem{Dosovitskiy2020}
Alexey Dosovitskiy, Lucas Beyer, Alexander Kolesnikov, Dirk Weissenborn,
  Xiaohua Zhai, Thomas Unterthiner, Mostafa Dehghani, Matthias Minderer, Georg
  Heigold, Sylvain Gelly, Jakob Uszkoreit, and Neil Houlsby.
\newblock An image is worth 16x16 words: Transformers for image recognition at
  scale.
\newblock {\em Proceedings of the International Conference on Learning
  Representations}, 2021.

\bibitem{Everingham15}
Mark Everingham, S.~M.~Ali Eslami, Luc~Van Gool, Chris K.~I. Williams, John
  Winn, and Andrew Zisserman.
\newblock The pascal visual object classes challenge: A retrospective.
\newblock {\em International Journal of Computer Vision}, 111(1):98--136,
  January 2015.

\bibitem{Fan2020}
Qi Fan, Wei Zhuo, Chi-Keung Tang, and Yu-Wing Tai.
\newblock Few-shot object detection with attention-rpn and multi-relation
  detector.
\newblock In {\em Proceedings of the IEEE Conference on Computer Vision and
  Pattern Recognition}, 2020.

\bibitem{Fan2021}
Zhibo Fan, Yuchen Ma, Zeming Li, and Jian Sun.
\newblock Generalized few-shot object detection without forgetting.
\newblock In {\em Proceedings of the IEEE Conference on Computer Vision and
  Pattern Recognition}, pages 4527--4536, 2021.

\bibitem{Girshick15}
Ross Girshick.
\newblock Fast {R-CNN}.
\newblock In {\em Proceedings of the International Conference on Computer
  Vision}, 2015.

\bibitem{Girshick13}
Ross Girshick, Jeff Donahue, Trevor Darrell, and Jitendra Malik.
\newblock Rich feature hierarchies for accurate object detection and semantic
  segmentation.
\newblock {\em arXiv preprint arXiv:1311.2524}, 2013.

\bibitem{Gupta2019}
Agrim Gupta, Piotr Dollar, and Ross Girshick.
\newblock {LVIS}: A dataset for large vocabulary instance segmentation.
\newblock In {\em Proceedings of the IEEE Conference on Computer Vision and
  Pattern Recognition}, 2019.

\bibitem{Han2021a}
Guangxing Han, Yicheng He, Shiyuan Huang, Jiawei Ma, and Shih-Fu Chang.
\newblock Query adaptive few-shot object detection with heterogeneous graph
  convolutional networks.
\newblock In {\em Proceedings of the International Conference on Computer
  Vision}, pages 3263--3272, October 2021.

\bibitem{He2020}
Kaiming He, Haoqi Fan, Yuxin Wu, Saining Xie, and Ross Girshick.
\newblock Momentum contrast for unsupervised visual representation learning.
\newblock In {\em Proceedings of the IEEE Conference on Computer Vision and
  Pattern Recognition}, pages 9729--9738, 2020.

\bibitem{Hinton2015}
Geoffrey Hinton, Oriol Vinyals, and Jeffrey Dean.
\newblock Distilling the knowledge in a neural network.
\newblock In {\em NeurIPS Deep Learning and Representation Learning Workshop},
  2015.

\bibitem{Hu2021}
Hanzhe Hu, Shuai Bai, Aoxue Li, Jinshi Cui, and Liwei Wang.
\newblock Dense relation distillation with context-aware aggregation for
  few-shot object detection.
\newblock In {\em Proceedings of the IEEE Conference on Computer Vision and
  Pattern Recognition}, pages 10185--10194, 2021.

\bibitem{Jeong19}
Jisoo Jeong, Seungeui Lee, Jeesoo Kim, and Nojun Kwak.
\newblock Consistency-based semi-supervised learning for object detection.
\newblock In {\em Advances in Neural Information Processing Systems}, 2019.

\bibitem{Kang2019}
Bingyi Kang, Zhuang Liu, Xin Wang, Fisher Yu, Jiashi Feng, and Trevor Darrell.
\newblock Few-shot object detection via feature reweighting.
\newblock In {\em Proceedings of the International Conference on Computer
  Vision}, pages 8420--8429, 2019.

\bibitem{Karlinsky2019}
Leonid Karlinsky, Joseph Shtok, Sivan Harary, Eli Schwartz, Amit Aides, Rogerio
  Feris, Raja Giryes, and Alex~M. Bronstein.
\newblock Repmet: Representative-based metric learning for classification and
  few-shot object detection.
\newblock In {\em Proceedings of the IEEE Conference on Computer Vision and
  Pattern Recognition}, 2019.

\bibitem{Khosla2020}
Prannay Khosla, Piotr Teterwak, Chen Wang, Aaron Sarna, Yonglong Tian, Phillip
  Isola, Aaron Maschinot, Ce Liu, and Dilip Krishnan.
\newblock Supervised contrastive learning.
\newblock {\em Advances in Neural Information Processing Systems}, 2020.

\bibitem{Law2018}
Hei Law and Jia Deng.
\newblock Cornernet: Detecting objects as paired keypoints.
\newblock In {\em Proceedings of the European Conference on Computer Vision},
  2018.

\bibitem{Li2021a}
Aoxue Li and Zhenguo Li.
\newblock Transformation invariant few-shot object detection.
\newblock In {\em Proceedings of the IEEE Conference on Computer Vision and
  Pattern Recognition}, pages 3094--3102, June 2021.

\bibitem{Li2021b}
Bohao Li, Boyu Yang, Chang Liu, Feng Liu, Rongrong Ji, and Qixiang Ye.
\newblock Beyond max-margin: Class margin equilibrium for few-shot object
  detection.
\newblock In {\em Proceedings of the IEEE Conference on Computer Vision and
  Pattern Recognition}, June 2021.

\bibitem{Li2019}
Yanghao Li, Yuntao Chen, Naiyan Wang, and Zhaoxiang Zhang.
\newblock Scale-aware trident networks for object detection.
\newblock In {\em Proceedings of the International Conference on Computer
  Vision}, 2019.

\bibitem{Li2021}
Yiting Li, Haiyue Zhu, Yu Cheng, Wenxin Wang, Chek~Sing Teo, Cheng Xiang,
  Prahlad Vadakkepat, and Tong~Heng Lee.
\newblock Few-shot object detection via classification refinement and
  distractor retreatment.
\newblock In {\em Proceedings of the IEEE Conference on Computer Vision and
  Pattern Recognition}, pages 15395--15403, June 2021.

\bibitem{Lin14}
Tsung{-}Yi Lin, Michael Maire, Serge~J. Belongie, Lubomir~D. Bourdev, Ross~B.
  Girshick, James Hays, Pietro Perona, Deva Ramanan, Piotr Doll{\'{a}}r, and
  C.~Lawrence Zitnick.
\newblock Microsoft coco: Common objects in context.
\newblock In {\em Proceedings of the European Conference on Computer Vision},
  2014.

\bibitem{Lin2017}
Tsung-Yi Lin, Piotr Dollar, Ross Girshick, Kaiming He, Bharath Hariharan, and
  Serge Belongie.
\newblock Feature pyramid networks for object detection.
\newblock In {\em Proceedings of the IEEE Conference on Computer Vision and
  Pattern Recognition}, 2017.

\bibitem{Lin2017a}
Tsung-Yi Lin, Priya Goyal, Ross Girshick, Kaiming He, and Piotr Doll{\'a}r.
\newblock Focal loss for dense object detection.
\newblock In {\em Proceedings of the International Conference on Computer
  Vision}, 2017.

\bibitem{Liu16}
Wei Liu, Dragomir Anguelov, Dumitru Erhan, Christian Szegedy, Scott Reed,
  Cheng-Yang Fu, and Alexander~C Berg.
\newblock Ssd: Single shot multibox detector.
\newblock In {\em Proceedings of the European Conference on Computer Vision},
  pages 21--37. Springer, 2016.

\bibitem{Liu21}
Yen-Cheng Liu, Chih-Yao Ma, Zijian He, Chia-Wen Kuo, Kan Chen, Peizhao Zhang,
  Bichen Wu, Zsolt Kira, and Peter Vajda.
\newblock Unbiased teacher for semi-supervised object detection.
\newblock In {\em Proceedings of the International Conference on Learning
  Representations}, 2021.

\bibitem{Liu2021}
Ze Liu, Yutong Lin, Yue Cao, Han Hu, Yixuan Wei, Zheng Zhang, Stephen Lin, and
  Baining Guo.
\newblock Swin transformer: Hierarchical vision transformer using shifted
  windows.
\newblock In {\em Proceedings of the International Conference on Computer
  Vision}, pages 10012--10022, October 2021.

\bibitem{Qiao2021a}
Limeng Qiao, Yuxuan Zhao, Zhiyuan Li, Xi Qiu, Jianan Wu, and Chi Zhang.
\newblock Defrcn: Decoupled faster r-cnn for few-shot object detection.
\newblock In {\em Proceedings of the International Conference on Computer
  Vision}, pages 8681--8690, October 2021.

\bibitem{Redmon16}
Joseph Redmon, Santosh Divvala, Ross Girshick, and Ali Farhadi.
\newblock You only look once: Unified, real-time object detection.
\newblock In {\em Proceedings of the IEEE Conference on Computer Vision and
  Pattern Recognition}, 2016.

\bibitem{Redmon2017}
Joseph Redmon and Ali Farhadi.
\newblock Yolo9000: Better, faster, stronger.
\newblock In {\em Proceedings of the IEEE Conference on Computer Vision and
  Pattern Recognition}, 2017.

\bibitem{Redmon2018}
Joseph Redmon and Ali Farhadi.
\newblock Yolov3: An incremental improvement.
\newblock {\em arXiv preprint arXiv:1804.02767}, 2018.

\bibitem{Ren16}
Shaoqing Ren, Kaiming He, Ross Girshick, and Jian Sun.
\newblock Faster {R-CNN}: Towards real-time object detection with region
  proposal networks.
\newblock In {\em Advances in Neural Information Processing Systems}, 2016.

\bibitem{Rosenberg05}
Chuck Rosenberg, Martial Hebert, and Henry Schneiderman.
\newblock Semi-supervised self-training of object detection models.
\newblock In {\em Winter Conference on Applications of Computer Vision}, 2005.

\bibitem{Scudder1965}
H. Scudder.
\newblock Probability of error of some adaptive pattern-recognition machines.
\newblock {\em IEEE Transactions on Information Theory}, pages 363--371, 1965.

\bibitem{Smith2017}
Linda~B. Smith and Lauren~K. Slone.
\newblock A developmental approach to machine learning?
\newblock {\em Frontiers in Psychology}, 2017.

\bibitem{Sohn20}
Kihyuk Sohn, Zizhao Zhang, Chun-Liang Li, Han Zhang, Chen-Yu Lee, and Tomas
  Pfister.
\newblock A simple semi-supervised learning framework for object detection.
\newblock {\em arXiv preprint arXiv:2005.04757}, 2020.

\bibitem{Sun2021}
Bo Sun, Banghuai Li, Shengcai Cai, Ye Yuan, and Chi Zhang.
\newblock Fsce: Few-shot object detection via contrastive proposal encoding.
\newblock In {\em Proceedings of the IEEE Conference on Computer Vision and
  Pattern Recognition}, June 2021.

\bibitem{Tan2020}
Mingxing Tan, Ruoming Pang, and Quoc~V Le.
\newblock Efficientdet: Scalable and efficient object detection.
\newblock In {\em Proceedings of the IEEE Conference on Computer Vision and
  Pattern Recognition}, 2020.

\bibitem{Tian2019}
Zhi Tian, Chunhua Shen, Hao Chen, and Tong He.
\newblock Fcos: Fully convolutional one-stage object detection.
\newblock In {\em Proceedings of the International Conference on Computer
  Vision}, 2019.

\bibitem{Wang2020}
Xin Wang, Thomas~E. Huang, Trevor Darrell, Joseph~E Gonzalez, and Fisher Yu.
\newblock Frustratingly simple few-shot object detection.
\newblock In {\em Proceedings of the International Conference on Machine
  Learning}, 2020.

\bibitem{Wang2019}
Yu-Xiong Wang, Deva Ramanan, and Martial Hebert.
\newblock Meta-learning to detect rare objects.
\newblock In {\em Proceedings of the International Conference on Computer
  Vision}, pages 9925--9934, 2019.

\bibitem{Wu2021a}
Aming Wu, Yahong Han, Linchao Zhu, and Yi Yang.
\newblock Universal-prototype enhancing for few-shot object detection.
\newblock In {\em Proceedings of the International Conference on Computer
  Vision}, pages 9567--9576, October 2021.

\bibitem{Wu2020a}
Jiaxi Wu, Songtao Liu, Di Huang, and Yunhong Wang.
\newblock Multi-scale positive sample refinement for few-shot object detection.
\newblock In {\em Proceedings of the European Conference on Computer Vision},
  2020.

\bibitem{Wu2019a}
Yuxin Wu, Alexander Kirillov, Francisco Massa, Wan-Yen Lo, and Ross Girshick.
\newblock Detectron2.
\newblock \url{https://github.com/facebookresearch/detectron2}, 2019.

\bibitem{Xiao2020}
Yang Xiao and Renaud Marlet.
\newblock Few-shot object detection and viewpoint estimation for objects in the
  wild.
\newblock In {\em Proceedings of the European Conference on Computer Vision},
  2020.

\bibitem{Xie2020a}
Qizhe Xie, Minh-Thang Luong, Eduard Hovy, and Quoc~V Le.
\newblock Self-training with noisy student improves imagenet classification.
\newblock In {\em Proceedings of the IEEE Conference on Computer Vision and
  Pattern Recognition}, 2020.

\bibitem{Xu2021}
Mengde Xu, Zheng Zhang, Han Hu, Jianfeng Wang, Lijuan Wang, Fangyun Wei, Xiang
  Bai, and Zicheng Liu.
\newblock End-to-end semi-supervised object detection with soft teacher.
\newblock 2021.

\bibitem{Yan2019}
Xiaopeng Yan, Ziliang Chen, Anni Xu, Xiaoxi Wang, Xiaodan Liang, and Liang Lin.
\newblock Meta r-cnn: Towards general solver for instance-level low-shot
  learning.
\newblock In {\em Proceedings of the International Conference on Computer
  Vision}, October 2019.

\bibitem{Yang2020}
Ze Yang, Yali Wang, Xianyu Chen, Jianzhuang Liu, and Yu Qiao.
\newblock Context-transformer: Tackling object confusion for few-shot
  detection.
\newblock 2020.

\bibitem{Zhou2019}
Xingyi Zhou, Dequan Wang, and Philipp Kr{\"a}henb{\"u}hl.
\newblock Objects as points.
\newblock {\em arXiv preprint arXiv:1904.07850}, 2019.

\bibitem{Zhu2021}
Chenchen Zhu, Fangyi Chen, Uzair Ahmed, Zhiqiang Shen, and Marios Savvides.
\newblock Semantic relation reasoning for shot-stable few-shot object
  detection.
\newblock In {\em Proceedings of the IEEE Conference on Computer Vision and
  Pattern Recognition}, pages 8782--8791, June 2021.

\bibitem{Zoph2020}
Barret Zoph, Golnaz Ghiasi, Tsung-Yi Lin, Yin Cui, Hanxiao Liu, Ekin~Dogus
  Cubuk, and Quoc Le.
\newblock Rethinking pre-training and self-training.
\newblock {\em Advances in Neural Information Processing Systems}, 2020.

\end{thebibliography}
}

\clearpage
% !TEX root = ../main.tex
\onecolumn

\begin{appendices}
{
  \hypersetup{linkcolor=black}
}

% !TEX root = ../supplementary_material.tex
\section{Supervision Collapse}
\label{app:supervision_collapse}
In Section 3 of the main paper, we experimentally demonstrate a form
of ``supervision collapse'' in few-shot object detection;
detector features trained {\em only on base class data} are naturally
biased against novel instance detection.  In this section we demonstrate two aspects of supervision collapse.

\begin{figure*}[p]
\centering
% \scriptsize
\includegraphics[width=\linewidth]{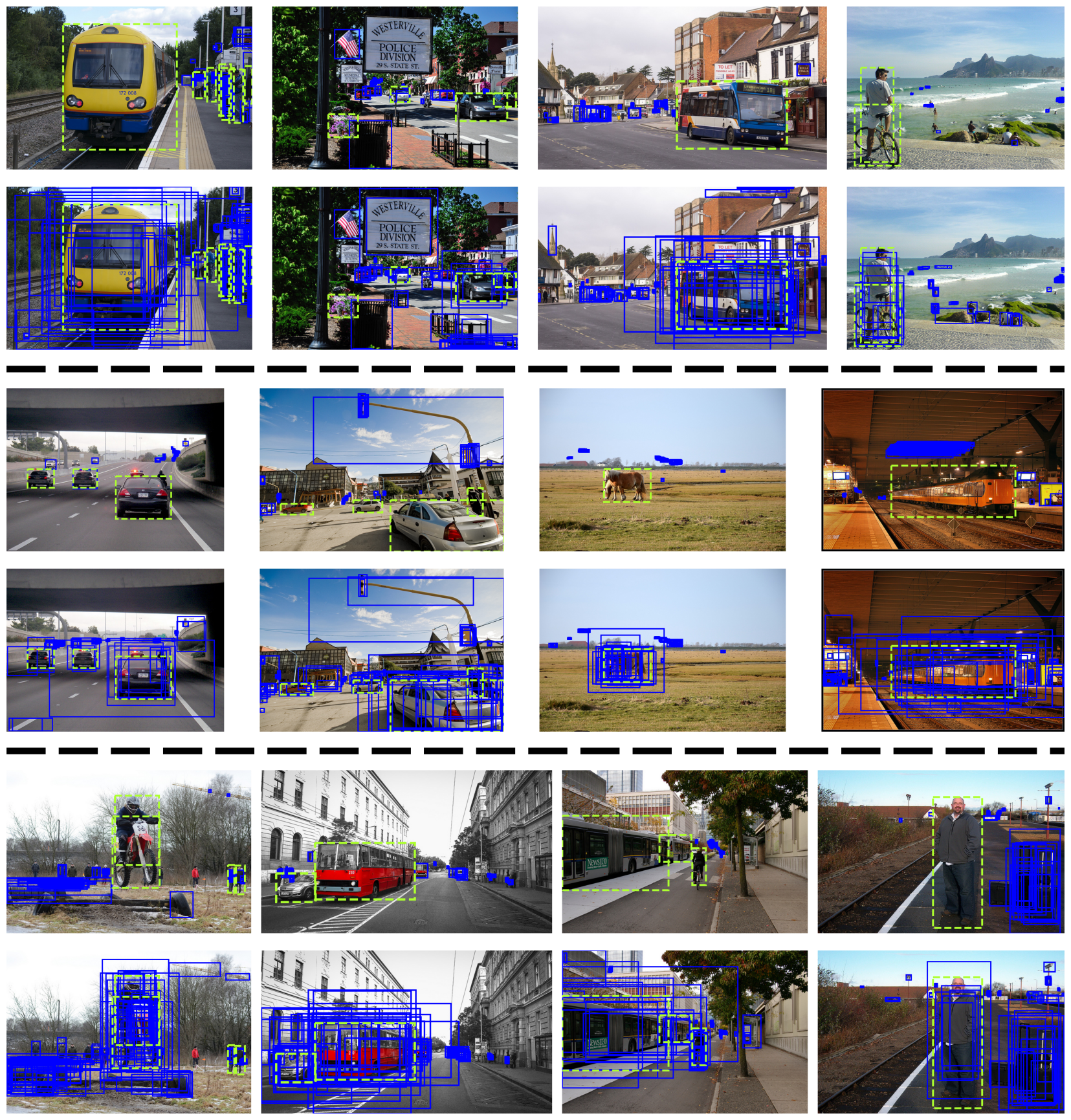}
% \vspace{-0.4cm}
\caption{ {\bf Supervision collapse for the  RPN.}
Top $100$ RPN proposals before and after {\bf Novel Training}.
In each pair of rows,
the upper row shows the top $100$ object proposals (blue) of an RPN trained {\em only on base class data}
and the lower row shows the top $100$ object proposals (blue) after RPN specific parameters ($\Phi_{\textsc{rpn}}$) have been trained on limited novel class data.
Groundtruth novel instances are shown in dashed lime.
We find an RPN trained only on base class data cannot propose many simple novel instances
and that training RPN specific parameters on limited novel class data leads to an
RPN which is substantially better at proposing object regions containing novel instances.}
% \hl{TODO} RPN trained only on Base vs RPN finetuned on novel data}
% \caption{\hl{TO DO}}
\label{fig:rpn_base_ft}
\end{figure*}

\subsection{On Generalisability of RPNs}

Figure~\ref{fig:rpn_base_ft} qualitatively illustrates the manifestation of ``supervision collapse''
in the Region Proposal Network (RPN) of a Faster R-CNN detector. 
We note that an RPN trained only with base class data struggles to propose boxes for the novel instances,
due to these novel instances having been treated as background during {\bf Base Training}.

We find that training RPN specific parameters ($2$ convolutional layers) on limited novel class data is adequate to drastically improve
the proposal of object regions containing novel instances.
This improvement is shown qualitatively in Figure~\ref{fig:rpn_base_ft} and quantitatively in Table 1 of the main paper.

\begin{figure*}[p]
\centering
% \scriptsize
\includegraphics[width=\linewidth]{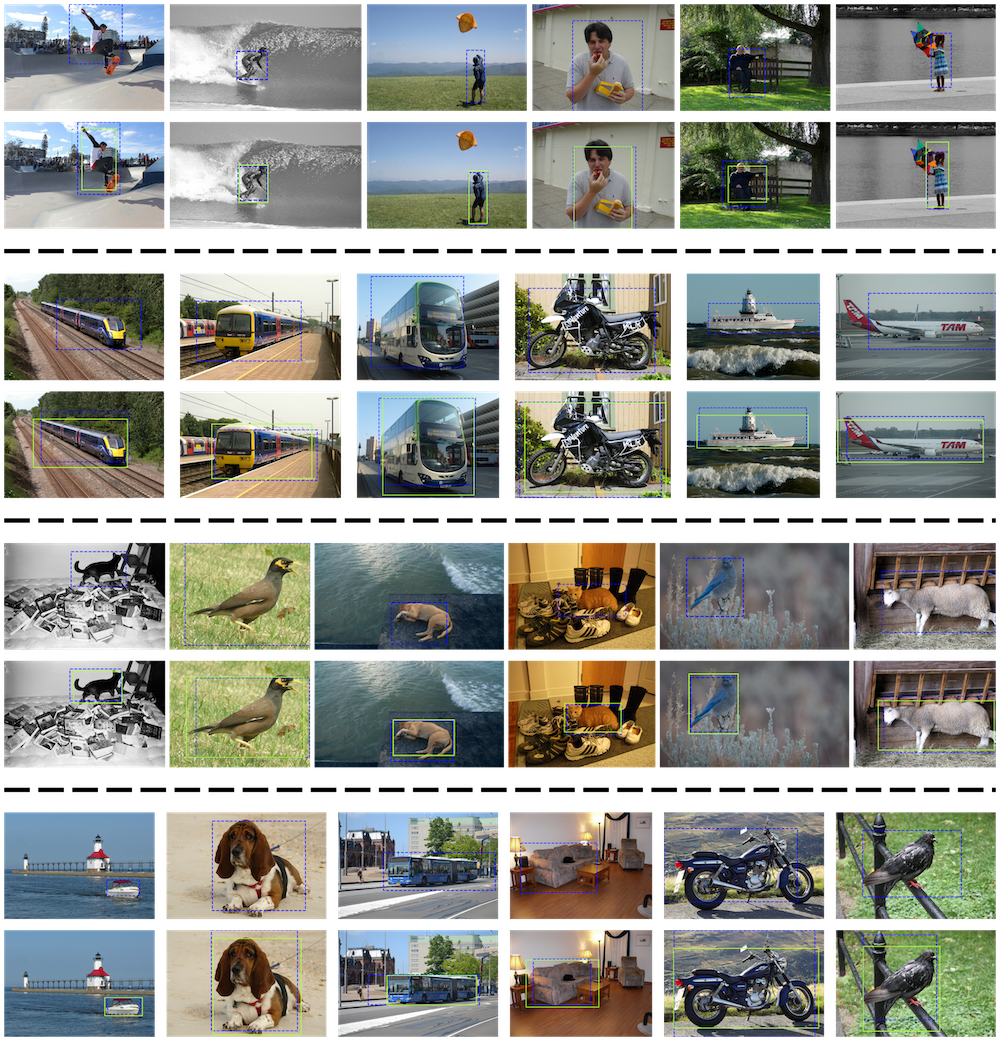}
% \vspace{-0.4cm}
\caption{
{\bf Supervision collapse for RoI classification.}
Comparing RoI classification between TFA~\cite{Wang2020}, where only $\Phi_{\textsc{cls}}$ is trained on limited novel class data,
with {\em our} {\bf Novel Training}, which additionally trains $\Phi_{\textsc{roi}}$ on limited novel class data, for cases where the object proposal is very good.
In each pair of rows,
the upper row shows the best object region (dashed blue) proposed by the RPN in TFA and the lack of foreground detection (false negative) from the second stage,
and the lower row shows the best object region (dashed blue) proposed by the RPN after {\em our} {\bf Novel Training}
and the correct foreground detection (solid lime). 
We find that TFA incorrectly classifies positive (with respect to novel classes) object regions (dashed blue) as background,
whereas {\em our} {\bf Novel Training} allows positive object regions (dashed blue) to be correctly classified as a novel instance (solid lime).}
% \hl{TODO} TFA incorrectly puts positive RPN proposals into background, our modifications to training prevent this form of supervision collapse}
% \caption{\hl{TO DO}}
\label{fig:tfa_supervision_collapse}
\end{figure*}

\subsection{On Transferability of Base Features}

Figure~\ref{fig:tfa_supervision_collapse} qualitatively demonstrates the other manifestation of ``supervision collapse'';
after training the second stage of Faster R-CNN as in~\cite{Wang2020}~\ie~only training the final classification layer on novel class data,
features in the second stage of the detector (which take RPN proposals as input),
incorrectly classify Regions-of-Interest (RoIs) containing novel instances as ``background'',
\ie~the detector features in the second stage do not
contain discriminative information to classify most novel instances, even in the cases where the object regions proposed by the RPN are very good.

We find that training parameters specific to the second stage of the detector, ($\Phi_{\textsc{cls}}$, $\Phi_{\textsc{roi}}$),
can moderately alleviate this form of ``supervision collapse'' such that the resultant detector can be used for Candidate Sourcing in our
Pseudo-Labelling method. Figure~\ref{fig:tfa_supervision_collapse} shows examples in which RoIs are correctly classified after {\em our}
{\bf Novel Training}.

\clearpage

% !TEX root = ../supplementary_material.tex
\section{Ignore Regions}
\label{app:ignore_regions}
\begin{figure*}[!htb]
\centering
% \scriptsize
\includegraphics[width=\linewidth]{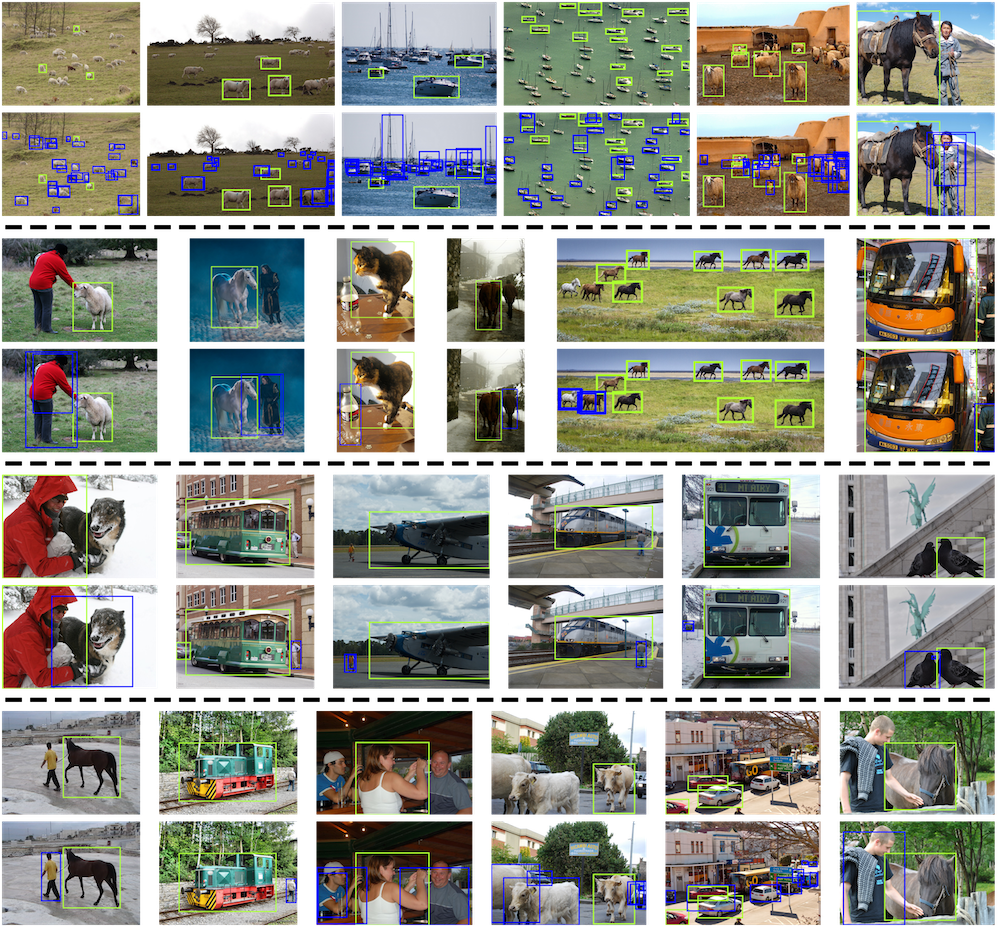}
% \vspace{-0.4cm}
\caption{
Comparing images in the training set with and without ignore regions.
In each pair of rows, the upper row shows images with pseudo-annotations found by our Pseudo-Labelling method (solid lime),
with many novel instances not covered,
the lower row shows the same images additionally showing the ignore regions (solid blue).
Using ignore regions covers many novel instances which are not found by our Pseudo-Labelling method,
preventing these novel instances being treated as background during end-to-end training.}
% \caption{\hl{TO DO}}
\label{fig:ignore_examples}
\end{figure*}

As mentioned in Section 5.2 of the main paper, 
we introduce the ``{\em ignore regions}'' during end-to-end training after making use of our Pseudo-Labelling method.
Figure~\ref{fig:ignore_examples} shows the impact of ignore regions.
% ** explain how ignore regions are used, i.e. where 
% they are and how they will influence the result.  **
During end-to-end training, we ensure no RoIs which have IoU$>0.5$ with an ignore region are sampled for training the second-stage of the detector;
RoIs which have IoU$>0.5$ with an ignore region are neither pushed into foreground (positive) or background (negative) in the second-stage of the detector.
Only with respect to the RPN during end-to-end training, ``{\em ignore regions}'' are treated as foreground~\ie~the RPN
after end-to-end training has learnt to propose regions-of-interest covering ignore regions.
Using ignore regions prevents the undiscovered novel instances
from being treated as background during end-to-end training~(as occurs during {\bf Base Training}) --
there are undiscovered novel instances because we use a score threshold of $0.8$ during Candidate Sourcing to yield high-precision pseudo-annotations.
We use all detections from our baseline detector, with novel class labels,
which are {\em unverified} after Label Verification as ignore regions
-- this includes detections with score less than $0.8$.

An ablation study on the utility of ignore regions is presented in Table~\ref{tab:ablation_ignore},
using the MS-COCO 30-shot object detection benchmark.
Using ignore regions improves performance by $4.4$, $3.5$ and $4.7$ on novel categories,
{\em i.e.}~nAP, nAP50 and nAP75, respectively, 
when compared to not using ignore regions.

% !TEX root = ../../supplementary_material.tex
\begin{table}[!htb]
\centering
\scriptsize
\begin{adjustbox}{width=.6\linewidth}
\setlength\tabcolsep{3.5pt}
\begin{tabular}{@{}l|c|ccc@{}}
\toprule
\multirow{2}{*}{Method} & \multirow{2}{*}{Ignore Regions?} & \multicolumn{3}{c}{Metrics} \\
                        &                                  & nAP     & nAP50   & nAP75   \\ \midrule
Ours (Pseudo-Labelling) & \xmark                           & 21.1    & 38.5    & 22.6    \\
Ours (Pseudo-Labelling) & \checkmark                       & 25.5    & 42.0    & 27.3    \\ \bottomrule
\end{tabular}
\end{adjustbox}

\caption{Ablation study on the utility of ignore regions.}
\label{tab:ablation_ignore}
\end{table}

\clearpage
% !TEX root = ../supplementary_material.tex
\section{More Qualitative Results}

% \input{sm_assets/sm_tables/ignore_regions_ablation.tex}
% \hl{To Do}
In this section we present more qualitative results from our Pseudo-Labelling method on the MS-COCO benchmark with $K$$=$$30$.

\subsection{Label Verification}
In Figure~\ref{fig:qual_verify}, we present additional qualitative results after
the {\em Label Verification} step of our Pseudo-Labelling method.
The first four rows in Figure~\ref{fig:qual_verify} show examples of Label Verification
(Section 4.2 of the main paper) in which the predicted class label from our baseline detector
(after {\bf Novel Training}) matches the class label from our $k$NN classifier.
These candidate detections are {\em verified} and used as inputs during {\em Box Correction} after which they are included as pseudo-annotations for end-to-end training.
The final two rows in Figure~\ref{fig:qual_correct} show examples of Label Verification
in which the predicted class label from our baseline detector {\em does not} match the class
label from our $k$NN classifier.
These candidate detections are {\em not verified} and rejected from our Pseudo-Labelling method.

\subsection{Box Correction}
\begin{figure*}[!htb]
\centering
% \scriptsize
\includegraphics[width=\linewidth]{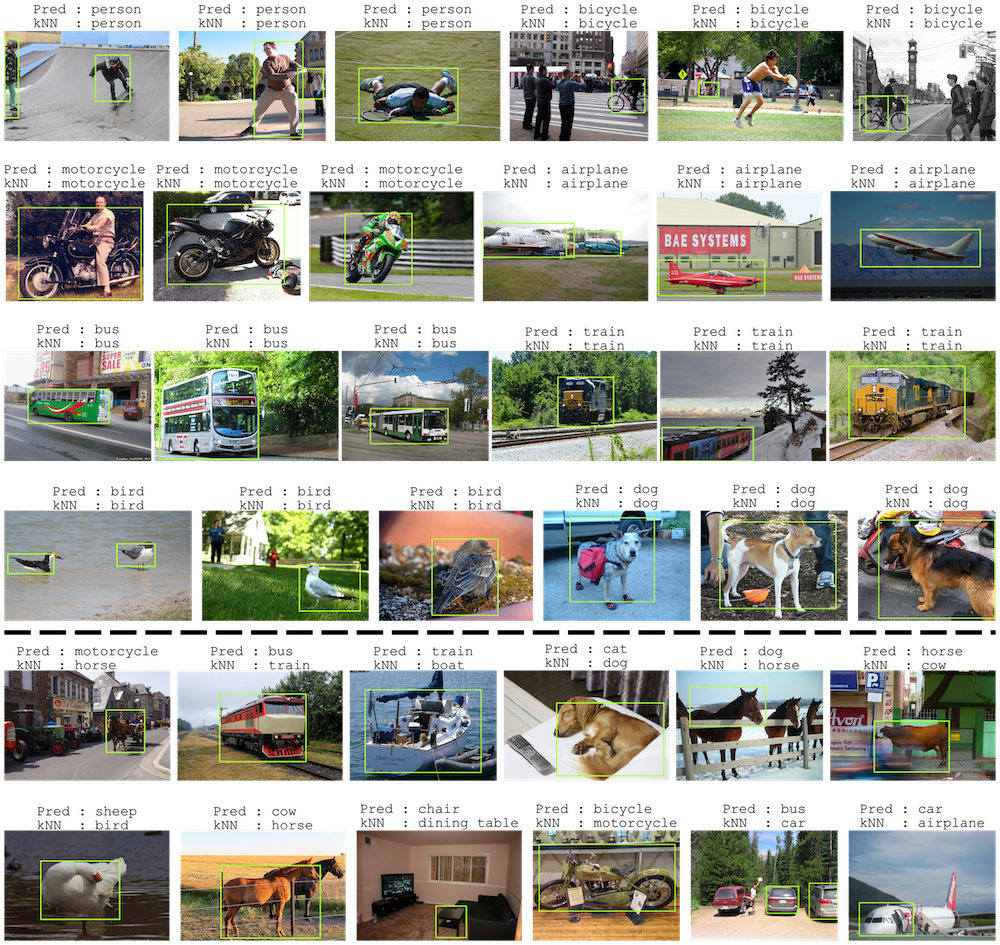}
% \vspace{-0.4cm}
\caption{More examples of {\em Label Verification}.
Top Four Rows: Candidate detections which {\em are verified} during {\em Label Verification};
the predicted class labels from our baseline detector and our $k$NN classifier match.
Bottom Two Rows: Candidate detections which {\em are not verified} (or rejected) during {\em Label Verification}; the predicted class labels from our baseline detector (false positive) and our $k$NN classifier {\em do not} match.}
% \caption{\hl{TO DO}}
\label{fig:qual_verify}
\end{figure*}

\begin{figure*}[!htb]
\centering
% \scriptsize
\includegraphics[width=\linewidth]{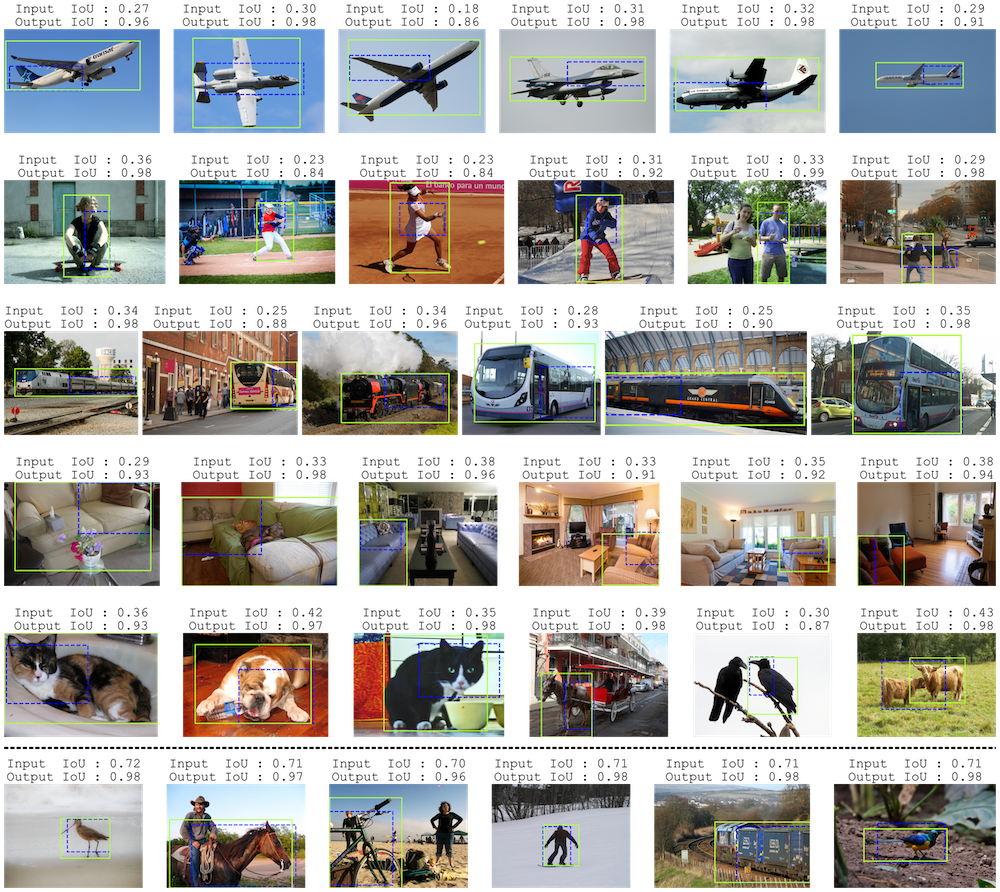}
% \vspace{-0.4cm}
\caption{More examples of {\em Box Correction}.
First Five Rows: Verified bounding boxes with poor quality/IoU with groundtruth (dashed blue) are drastically improved (solid lime) during {\em Box Correction}.
Bottom Row: Verified bounding boxes with acceptable quality/IoU with groundtruth (dashed blue) are further improved (solid lime) during {\em Box Correction}.}
% \caption{\hl{TO DO}}
\label{fig:qual_correct}
\end{figure*}

In Figure~\ref{fig:qual_correct}, we present additional qualitative results after
the {\em Box Correction} step of our Pseudo-Labelling method.
The first five rows in Figure~\ref{fig:qual_correct} show examples of Box Correction
(Section 4.3 of the main paper) in which the bounding box coordinates of {\em verified} candidate detections are poor (dashed blue)~\ie~they have low intersection-over-union (IoU) with the groundtruth instance.
During {\em Box Correction} these poor bounding boxes are drastically improved to near perfect bounding boxes (solid lime).
The bottom row in Figure~\ref{fig:qual_verify} show examples of Box Correction
in which the bounding box coordinates of the verified candidate detections (dashed blue) have acceptable 
IoU with groundtruth instances.
During {\em Box Correction} these acceptable bounding boxes are further improved to near
perfect bounding boxes (solid lime).

% \weidi{use subsection to describe what will be shown in Figure 4, 5}

\clearpage
% !TEX root = ../supplementary_material.tex
\section{Quantitative Results}

In this section we present quantitative results of each step of our Pseudo-Labelling method on the MS-COCO benchmark with $K$$=$$30$.
We show precision-recall curves using the various IoU criteria used in MS-COCO.

\begin{figure*}[!htb]
\centering
% \scriptsize
\includegraphics[width=\linewidth]{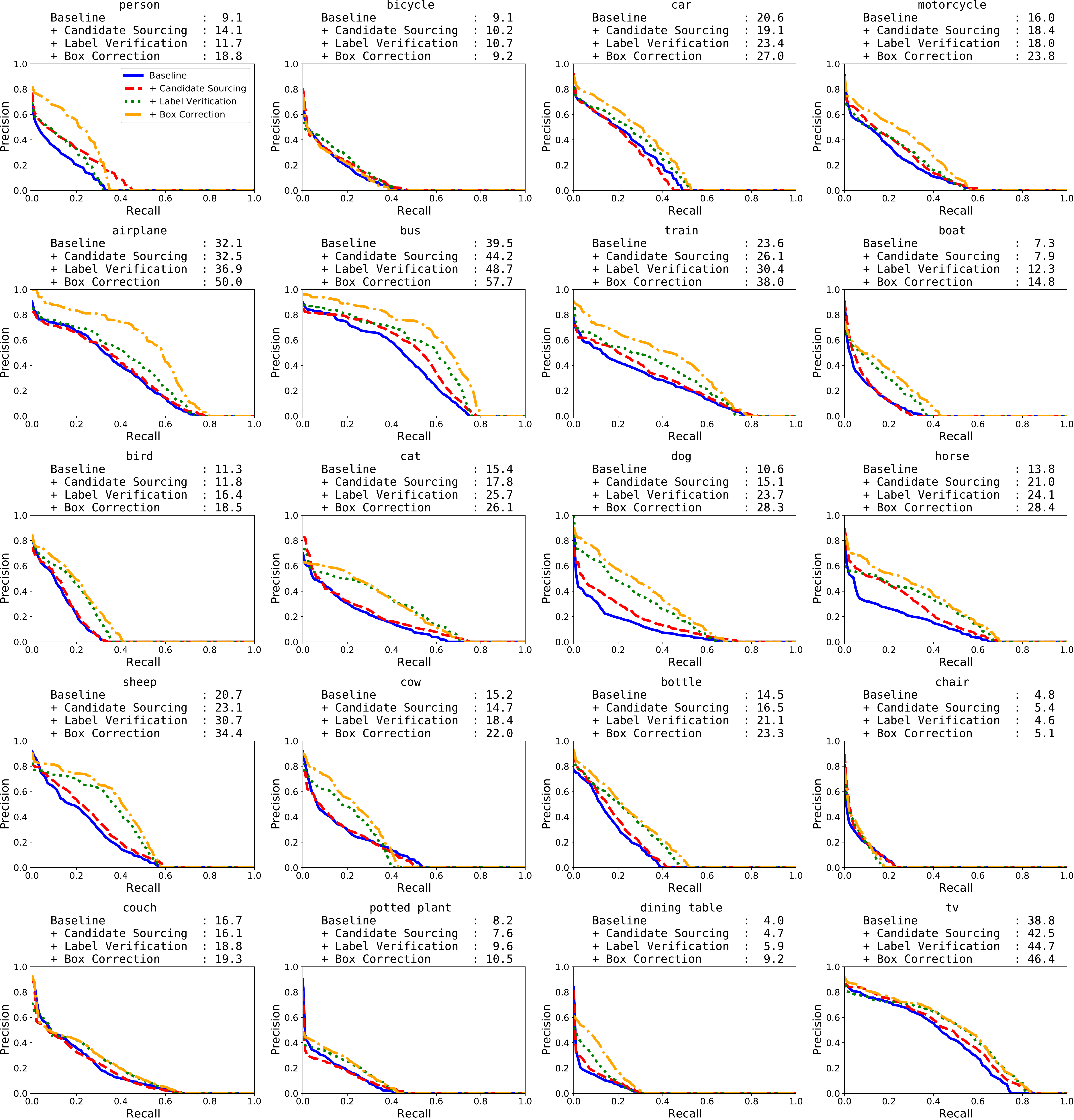}
% \vspace{-0.4cm}
\caption{
{\bf Precision-Recall curves for novel classes with IoU=0.5:0.95.}
Baseline curves are shown in solid blue.
Curves are shown after adding Candidate Sourcing, Label Verification and Box Correction
in dashed red, dotted green and dash-dotted orange, respectively.
The nAP value for each step of our method is shown in the title of each subplot.
}
\label{fig:quant_nAP}
\end{figure*}

\begin{figure*}[!htb]
\centering
% \scriptsize
\includegraphics[width=\linewidth]{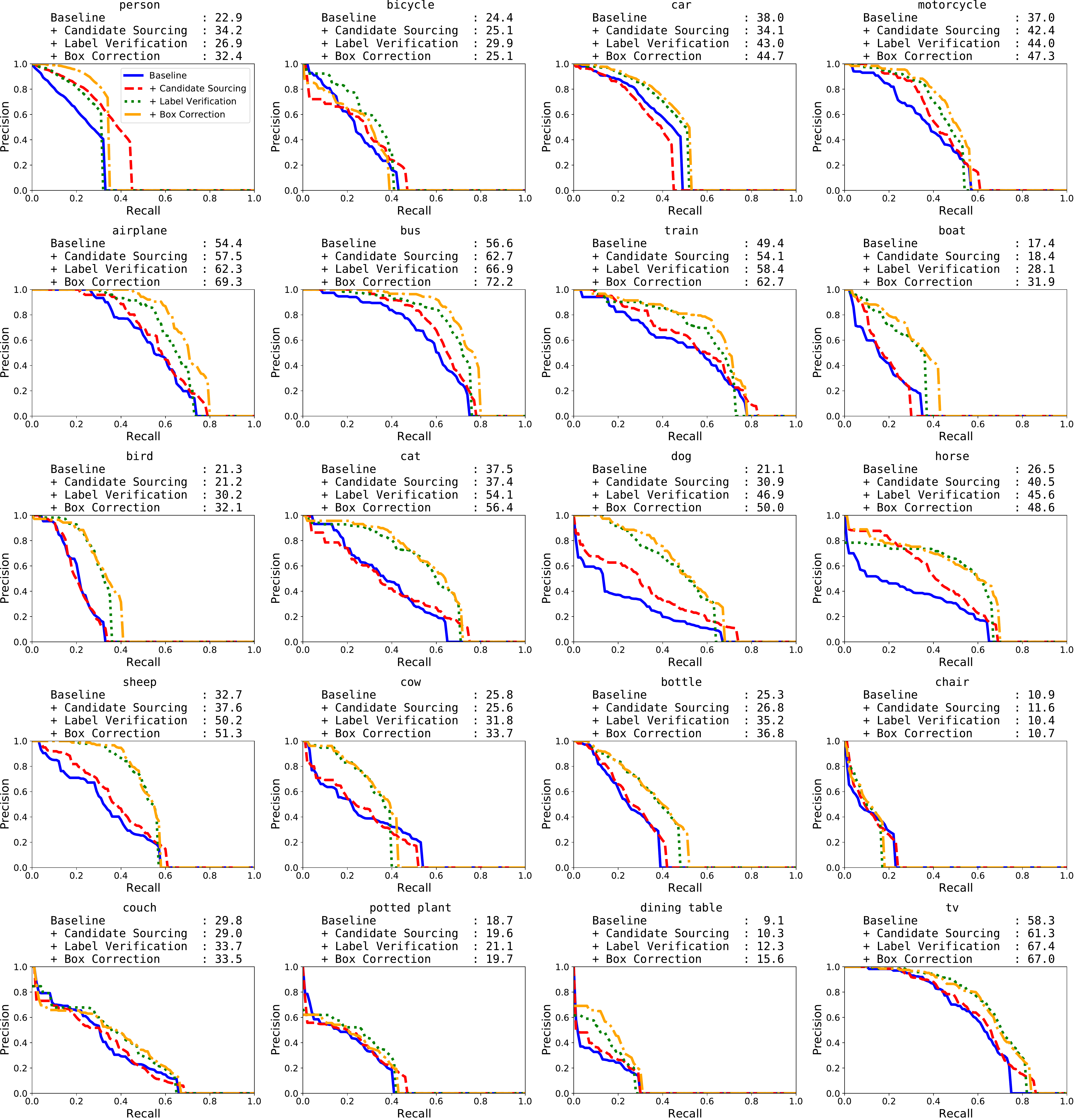}
% \vspace{-0.4cm}
\caption{
{\bf Precision-Recall curves for novel classes with IoU=0.5.}
Baseline curves are shown in solid blue.
Curves are shown after adding Candidate Sourcing, Label Verification and Box Correction
in dashed red, dotted green and dash-dotted orange, respectively.
The nAP50 value for each step of our method is shown in the title of each subplot.
}
\label{fig:quant_nAP50}
\end{figure*}

\begin{figure*}[!htb]
\centering
% \scriptsize
\includegraphics[width=\linewidth]{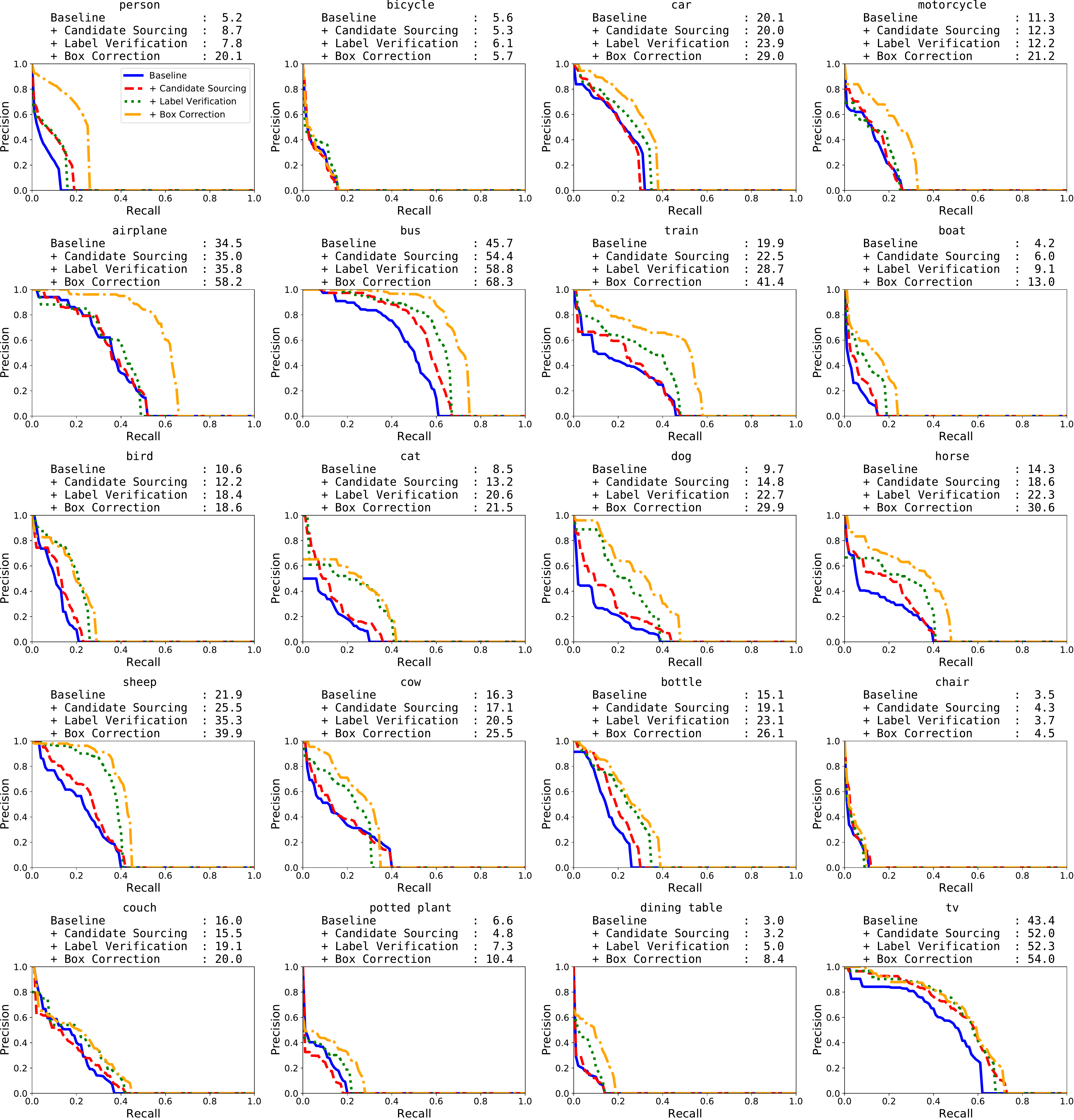}
% \vspace{-0.4cm}
\caption{
{\bf Precision-Recall curves for novel classes with IoU=0.75.}
Baseline curves are shown in solid blue.
Curves are shown after adding Candidate Sourcing, Label Verification and Box Correction
in dashed red, dotted green and dash-dotted orange, respectively.
The nAP75 value for each step of our method is shown in the title of each subplot.
}
\label{fig:quant_nAP75}
\end{figure*}

Precision-recall curves when using the IoU=0.5, IoU=0.75 and IoU=0.5:0.95 MS-COCO criteria are shown in
Figure~{\ref{fig:quant_nAP},\ref{fig:quant_nAP50},\ref{fig:quant_nAP75}}, respectively.

For the vast majority of novel classes, adding our Pseudo-Labelling method improves performance over our baseline detector performance,
regardless of which MS-COCO criteria is used.
However, we note that the effect of each step is not uniform across different novel classes and metrics.
For example, considering the \texttt{person} class, using Candidate Sourcing only is the best performing model when using IoU=0.5 due to higher recall,
but when using IoU=0.75 and IoU=0.5:0.95 our full Pseudo-Labelling method performs best.

In terms of IoU=0.5:0.95, which is the main MS-COCO criterion,
Figure~\ref{fig:quant_nAP} shows that different classes see the largest performance boost from different steps in our Pseudo-Labelling process.
For example, the largest performance boost comes from Label Verification from some classes~\eg~\texttt{bird}, \texttt{sheep}, \texttt{cat};
these classes suffer from many false positives with respect to class labels after Candidate Sourcing,
which are removed during Label Verification yielding better performance.
On the other hand, for other classes the largest performance boost comes from Box Correction~\eg~\texttt{airplane}, \texttt{train}, \texttt{person};
these classes suffer from many false positives with respect to bounding box coordinates after Candidate Sourcing,
which are corrected during Box Correction yielding better performance.

We note that in some rare cases an individual Pseudo-Labelling step may harm performance,~\eg~adding
Label Verification for \texttt{person} reduces performance by $2.4$nAP compared to using Candidate Sourcing only,
and adding Box Correction for \texttt{bicycle} reduces performance by $1.5$nAP compared to using Candidate Sourcing and Label Verification only.

Moreover, we note that our Pseudo-Labelling method yields negligible performance improvements for the \texttt{bicycle} and \texttt{chair} classes yielding
only a $0.1$ and $0.3$nAP boost, respectively, over our baseline detector.

% \weidi{use subsection to describe what will be shown in Figure 4, 5}

\clearpage
% !TEX root = ../supplementary_material.tex
\section{Limitations}
\label{app:limitations}
There are two main limitations of our work:
{\em first}, our work follows prior work and performs benchmark experiments on
MS-COCO~\cite{Lin14} and PASCAL VOC~\cite{Everingham15},
which contain $80$ and $20$ classes, respectively.
In future work, we will benchmark the idea on LVIS~\cite{Gupta2019},
which contains $1230$ classes, and better characterise few-shot object detection methods.
{\em Second}, in this work, we only perform one iteration of pseudo-labelling with $k$NN classifier.
Given our pseudo-annotations are noisy, one can make use of noisy label learning
to effectively incorporate our pseudo-annotations as training data for training more powerful classifiers in the Label Verification step and obtaining more pseudo-annotations.

\section{Ethical Concerns}
\label{app:ethical_concerns}
Our proposed method allows object detectors to detect novel categories with limited
access to groundtruth annotations.
Moreover, our work contains a Pseudo-Labelling method which enables object detectors to 
effectively re-train on (noisy) pseudo-annotations.
Our work could be applied to numerous areas~\eg~public surveillance and autonomous driving.
While not our intention, our method could be used by agents with malicious purposes for
detection with limited data.
We think robust public debate and suitable regulations can reduce the risk of malicious
applications of our work.

\end{appendices}

% \includepdf[pages={1-last}]{supplementary_material/supplementary_material.pdf}

\end{document}